\documentclass[sigconf,nonacm]{acmart}
\AtBeginDocument{%
  }

\setcopyright{acmlicensed}
\copyrightyear{2026}
\acmYear{2026}
\acmDOI{XXXXXXX.XXXXXXX}
\acmISBN{978-1-4503-XXXX-X/2018/06}

\usepackage{booktabs}
\usepackage{multirow}
\usepackage{graphicx}

\usepackage{graphicx}
\usepackage{subcaption}

\usepackage[table]{xcolor}

\renewcommand\footnotetextcopyrightpermission[1]{}
\begin{document}

%%
%% The "title" command has an optional parameter,
%% allowing the author to define a "short title" to be used in page headers.
\title{Bridging the RGB-IR Gap: Consensus and Discrepancy Modeling for Text-Guided Multispectral Detection}

%%
%% The "author" command and its associated commands are used to define
%% the authors and their affiliations.
%% Of note is the shared affiliation of the first two authors, and the
%% "authornote" and "authornotemark" commands
%% used to denote shared contribution to the research.

\author{Jiaqi Wu}
\affiliation{%
  \institution{Department of Automation, Tsinghua University }
  \city{BeiJing}
  \country{China}
}

\author{Zhen Wang}
\affiliation{%
  \institution{School of Artificial Intelligence , China University of Mining Technology - Beijing }
  \city{BeiJing}
  \country{China}
}
\author{Enhao Huang}
\affiliation{%
  \institution{State Key Laboratory of Blockchain and Data Security, Zhejiang University}
  \city{HangZhou}
  \country{China}
}
\author{Kangqing Shen}
\affiliation{%
  \institution{Department of Automation, Tsinghua University }
  \city{BeiJing}
  \country{China}
}
\author{Yulin Wang}
\affiliation{%
  \institution{Tsinghua University}
  \city{BeiJing}
  \country{China}
}
\author{Yang Yue}
\affiliation{%
  \institution{Department of Automation, Tsinghua University}
  \city{BeiJing}
  \country{China}
}
\author{Yifan Pu}
\affiliation{%
  \institution{
Tsinghua University }
  \city{BeiJing}
  \country{China}
}
\author{Gao Huang}
\affiliation{%
  \institution{Department of Automation, Tsinghua University }
  \city{BeiJing}
  \country{China}
}
\renewcommand{\shortauthors}{}

%%
%% The abstract is a short summary of the work to be presented in the
%% article.
\begin{abstract}
Text-guided multispectral object detection uses text semantics to guide semantic-aware cross-modal interaction between RGB and IR for more robust perception. However, notable limitations remain: (1) existing methods often use text only as an auxiliary semantic enhancement signal, without exploiting its guiding role to bridge the inherent granularity asymmetry between RGB and IR; and (2) conventional data-driven attention-based fusion tends to emphasize stable consensus while overlooking potentially valuable cross-modal discrepancies. To address these issues, we propose a semantic bridge fusion framework with bi-support modeling for multispectral object detection. Specifically, text is used as a shared semantic bridge to align RGB and IR responses under a unified category condition, while the recalibrated thermal semantic prior is projected onto the RGB branch for semantic-level mapping fusion. We further formulate RGB-IR interaction evidence into the regular consensus support and the complementary discrepancy support that contains potentially discriminative cues, and introduce them into fusion via dynamic recalibration as a structured inductive bias. In addition, we design a bidirectional semantic alignment module for closed-loop vision-text guidance enhancement. Extensive experiments demonstrate the effectiveness of the proposed fusion framework and its superior detection performance on multispectral benchmarks. Code is available at \url{https://github.com/zhenwang5372/Bridging-RGB-IR-Gap}.
\end{abstract}
% 

%%
%% The code below is generated by the tool at http://dl.acm.org/ccs.cfm.
%% Please copy and paste the code instead of the example below.
%%

% \ccsdesc[500]{Do Not Use This Code~Generate the Correct Terms for Your Paper}
% \ccsdesc[300]{Do Not Use This Code~Generate the Correct Terms for Your Paper}
% \ccsdesc{Do Not Use This Code~Generate the Correct Terms for Your Paper}
% \ccsdesc[100]{Do Not Use This Code~Generate the Correct Terms for Your Paper}

%%
%% Keywords. The author(s) should pick words that accurately describe
%% the work being presented. Separate the keywords with commas.

%\keywords{Multispectral object detection, semantic multi-modal fusion, cross-modal semantic modeling}

%% A "teaser" image appears between the author and affiliation
%% information and the body of the document, and typically spans the
%% page.
% \begin{teaserfigure}
%   \includegraphics[width=\textwidth]{samples/Images/Overview1.pdf}
%   \caption{Seattle Mariners at Spring Training, 2010.}
%   \Description{Enjoying the baseball game from the third-base
%   seats. Ichiro Suzuki preparing to bat.}
%   \label{fig:teaser}
% \end{teaserfigure}

\begin{teaserfigure}
  \includegraphics[width=\textwidth]{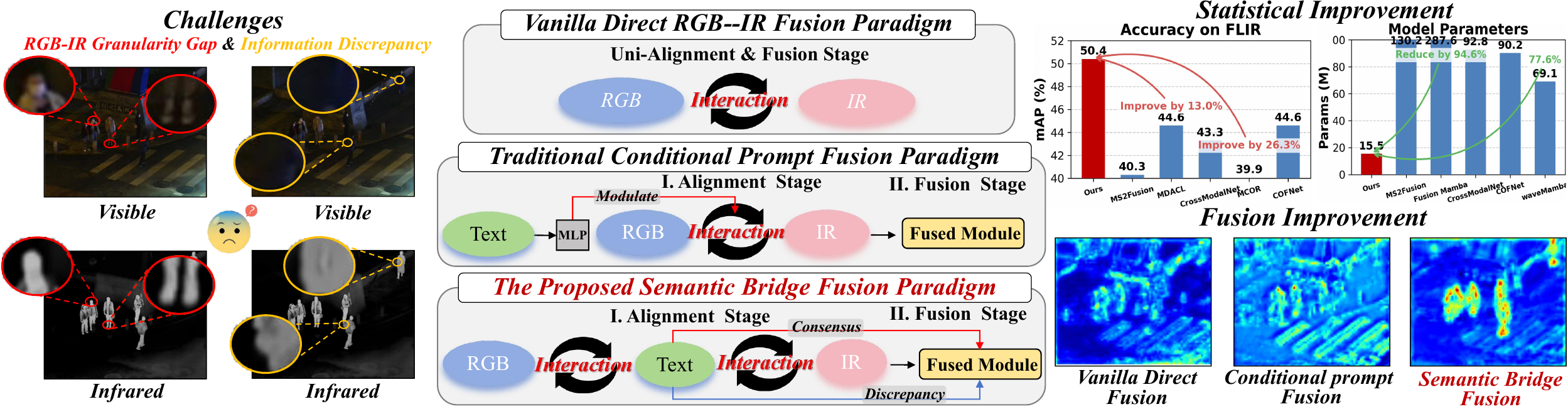}
  % \vspace{-0.7cm}  
  \caption{Comparison of different fusion paradigms. Unlike vanilla direct RGB--IR fusion and conditional prompt fusion, our semantic-bridge fusion explicitly exploits textual semantics to enable semantic-level RGB--IR interaction, while introducing consensus--discrepancy support modeling to facilitate fusion. This leads to more salient target responses in the fused features, achieving the best detection accuracy with significantly fewer parameters.}
  \Description{An overview figure comparing different text-guided RGB--IR fusion paradigms, including conventional RGB--IR fusion, conditional prompt fusion, and the proposed semantic-bridge fusion. The figure also shows consensus and discrepancy support patterns, as well as fused results in which the target regions are more salient under the proposed method.}
  \label{fig:teaser}
\end{teaserfigure}

% \received{20 February 2007}
% \received[revised]{12 March 2009}
% \received[accepted]{5 June 2009}

%%
%% This command processes the author and affiliation and title
%% information and builds the first part of the formatted document.
\maketitle

% \begin{figure}[t!]
%   \centering
%   \subfloat[RGB-IR Fusion]{
%     \includegraphics[width=0.15\textwidth,height=2.4cm]{samples/Images/rgb-ir-CA.png}
%   }
%   \subfloat[Text Prompt Fusion]{
%     \includegraphics[width=0.15\textwidth,height=2.4cm]{samples/Images/textinfusion.png}
%   }
%   \subfloat[Our Semantic Bridge Fusion]{
%     \includegraphics[width=0.15\textwidth,height=2.4cm]{samples/Images/semanticguidefusion.png}
%   }
%   \vspace{-0.4cm}
%   \caption{Comparison of three representative fusion paradigms for multispectral object detection: conventional RGB-IR fusion, conditional prompt-based fusion, and our semantic-bridge-guided fusion.}
%   \Description{Three images arranged in a single row across two columns, illustrating RGB-IR fusion, conditional prompt fusion, and semantic bridge fusion, respectively.}
%   \label{fig:three_fusion_paradigms}
% \end{figure}

\section{Introduction}
% Multispectral object detection jointly exploits RGB and IR information to achieve more robust target perception in complex environments \cite{liu2022target,cao2021handcrafted}. Existing RGB–IR methods mostly rely on static fusion at the pixel, channel, or feature level\cite{li2019illumination,zhang2023visible}. However, such paradigms implicitly assume that responses from different modalities can be integrated in a relatively uniform manner. In practice, RGB is more effective at capturing texture, edges, and appearance details, whereas IR is more sensitive to regional thermal responses and target existence cues\cite{liu2024infrared}. The two modalities are therefore inherently asymmetric in both information granularity and representation mechanism\cite{zhou2020improving,kim2021uncertainty}. Consequently, the key to RGB–IR fusion lies not in “fusing more information,” but in determining how to achieve more reasonable semantic-level interaction under unified target semantics. 

% As illustrated in Fig. \ref{fig:three_fusion_paradigms}(a), without shared semantic constraints, cross-modal responses are more likely to remain at the level of coarse-grained correlation. In response, recent studies have begun to introduce text into RGB–IR detection to provide category-level semantic information\cite{wang2026fdfusion,yi2024text}. Nevertheless, most of them only use text as an additional semantic enhancement signal or conditional prompt, as shown in Fig. \ref{fig:three_fusion_paradigms}(b) with still limited effectiveness.

Multispectral object detection jointly exploits RGB and IR information to achieve more robust target perception in complex environments \cite{liu2022target,cao2021handcrafted}. Existing RGB--IR methods mostly rely on static fusion at the pixel, channel, or feature level \cite{li2019illumination,zhang2023visible}. However, such paradigms implicitly assume that responses from different modalities can be integrated in a relatively uniform manner. In practice, RGB is more effective at capturing texture, edges, and appearance details, whereas IR is more sensitive to regional thermal responses and target existence cues \cite{liu2024infrared}. The two modalities are therefore inherently asymmetric in both information granularity (as shown in Fig.~\ref{fig:teaser}-\textit{Challenges}-\textit{RGB-IR Granularity Gap}) and representation mechanism \cite{zhou2020improving,kim2021uncertainty}. Consequently, the key to RGB--IR fusion lies not in simply fusing more information, but in establishing more reasonable semantic-level interaction under unified target semantics. As shown in Fig.~\ref{fig:teaser} (\textit{Vanilla Direct RGB-IR Fusion}), if RGB and IR are directly fused, cross-modal responses tend to remain at a coarse-grained level, making it difficult to distinguish foreground targets from background clutter. To address this issue, recent studies have begun to introduce text into RGB--IR detection to provide category-level semantic information \cite{wang2026fdfusion,yi2024text}. Nevertheless, as shown in Fig.~\ref{fig:teaser} (\textit{Traditional Conditional Prompt Fusion}), such methods still treat text merely as an additional semantic enhancement signal or conditional prompt to modulate RGB--IR fusion\cite{zhang2025omnifuse, Tang2025ControlFusion,SLGNets}. Since the semantic property of text is not explicitly exploited to organize cross-modal interaction, target perception remains limited and is still easily disturbed by background responses.

Based on the above observations, we argue that, in multispectral detection, text is more appropriately regarded as a shared semantic bridge \cite{radford2021learning}, as shown in Fig.~\ref{fig:teaser} (\textit{The Proposed Semantic Bridge Fusion-I. Alignment Stage}). On the one hand, it provides a unified category-semantic condition for RGB and IR, thereby establishing the basis for target-semantic alignment between the two modalities. On the other hand, it further promotes cross-modal mapping and fusion through semantic guidance, elevating cross-modal interaction from static feature aggregation to semantic-level support modeling. 

\begin{figure}[h!]
  \centering
  \subfloat[Consensus activation.]{
    \includegraphics[width=0.23\textwidth,height=3.7cm]{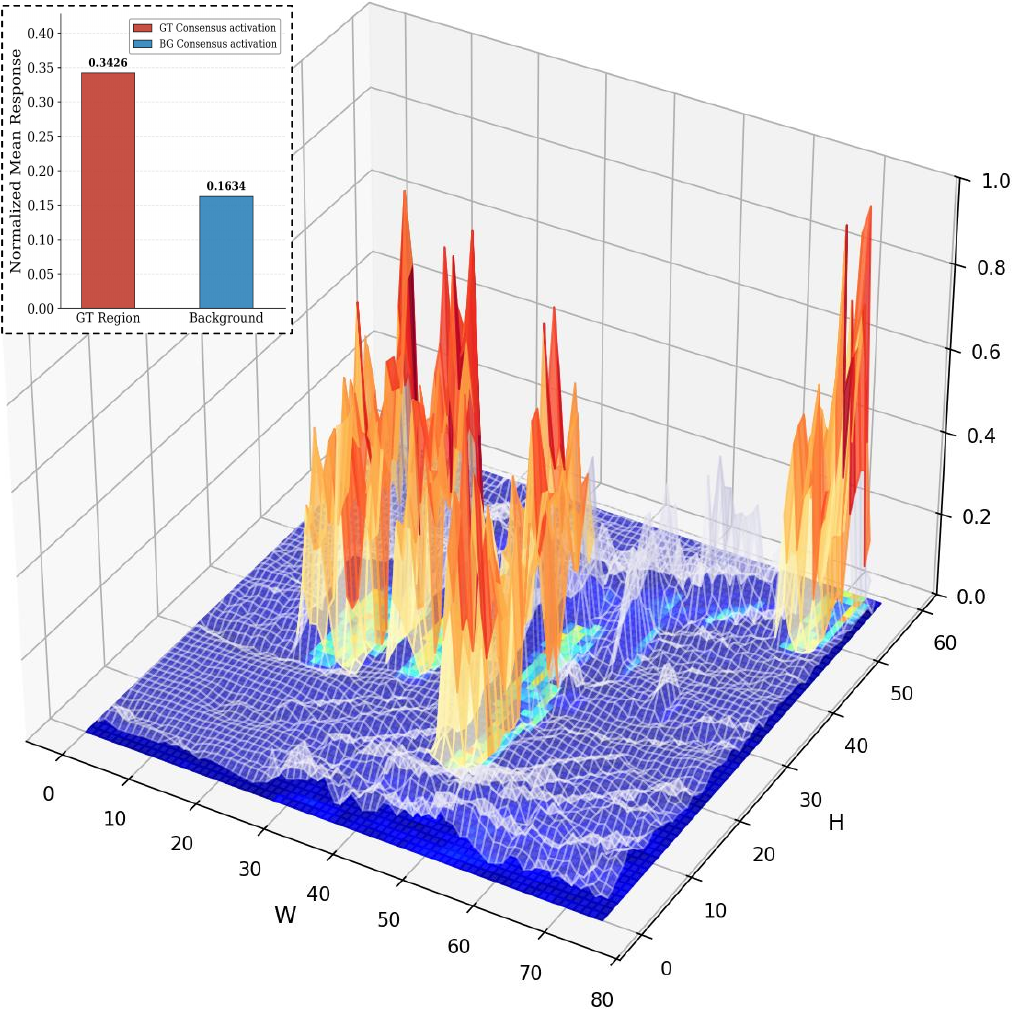}
  }
  \subfloat[Consensus\&discrepancy activation.]{
    \includegraphics[width=0.23\textwidth,height=3.7cm]{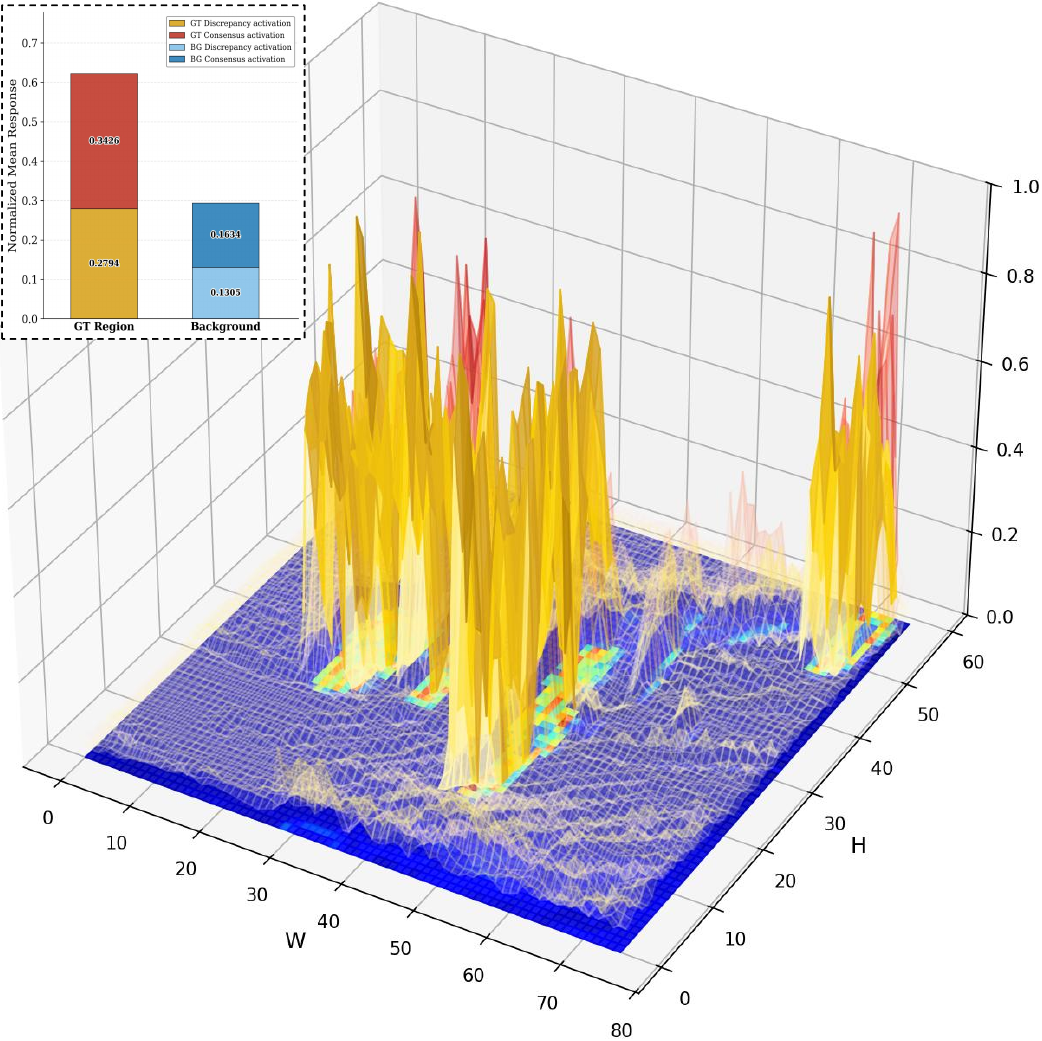}
  }
  \vspace{-0.4cm}
\caption{Visualization of consensus–discrepancy activation
under dark conditions. The peak-distribution and GT-box statistics indicate that consensus constitutes an important support pattern in (a). In (b), discrepancy is further incorporated based on (a), thereby providing a more complete characterization of target perception.}
  \Description{Two activation maps under dark conditions, showing discrepancy support on the left and consensus support on the right. Additional rectangular plots compare activation statistics in GT boxes and background regions.}
  \label{fig:bi_support_dark}
\end{figure}

However, text-guided cross-modal semantic consensus---i.e., the spatial regions jointly activated by RGB and IR under the same textual semantics---is still insufficient to fully characterize multispectral fusion, due to the inherent information discrepancy between the two modalities (see Fig.~\ref{fig:teaser}, \textit{Challenges}--\textit{Information Discrepancy}). Taking the representative degraded scenario in Fig.~\ref{fig:teaser} (\textit{Challenges}) as an example, we further visualize the support patterns in Fig.~\ref{fig:bi_support_dark}. Although consensus activations cover a large portion of the GT regions and already provide useful target support, as shown in Fig.~\ref{fig:bi_support_dark}(a), the semantic regions in degraded scenes still exhibit substantial fine-grained differences. Importantly, discrepancy support should not be regarded as erroneous response; rather, it reflects an inconsistent support state that may still preserve valid target evidence. Fig.~\ref{fig:bi_support_dark}(b) characterizes this phenomenon, showing that a considerable amount of discrepancy activation is also distributed within the GT box, and that consensus and discrepancy together form complementary support patterns for target perception. Moreover, since multispectral detection is often deployed in degraded scenarios, such discrepancy is common and should not be ignored.

% \begin{figure}[t!]
%   \centering
%   \includegraphics[width=0.15\textwidth,height=2.4cm]{samples/Images/rgb-ir-CA.png}
%   \includegraphics[width=0.15\textwidth,height=2.4cm]{samples/Images/textinfusion.png}
%   \includegraphics[width=0.15\textwidth,height=2.4cm]{samples/Images/semanticguidefusion.png}
%   \caption{Visualization results of five representative samples.}
%   \Description{Five images arranged in a single row across two columns, showing representative multispectral visual examples.}
%   \label{fig:five_images_row}
% \end{figure}

% \begin{figure}[h!]
%   \centering
% \includegraphics[width=0.23\textwidth]{samples/Images/sample-descrepancy.png}
%   \includegraphics[width=0.23\textwidth]{samples/Images/sample-Consensus.png}
%   \caption{Visualization results of five representative samples.}
%   \Description{Five images arranged in a single row across two columns, showing representative multispectral visual examples.}
%   \label{fig:five_images_row}
% \end{figure}

Conventional data-driven end-to-end attention-based fusion paradigms tend to reinforce consensus activations \cite{qingyun2021cross}, and are therefore prone to overlooking informative cues embedded in discrepancy activation. Accordingly, we explicitly characterize the support pattern of the two modalities for the same target semantics by two distinct states: \textbf{consensus support}, in which RGB and IR jointly support the target semantics, and \textbf{discrepancy support}, which arises from modality-adaptation differences and asynchronous information loss. At the same time, discrepancy support should not be preserved unconditionally, since it may also contain spurious responses. Therefore, the proper strategy is neither to simply discard discrepancy nor to directly amplify it, but to exploit it through \textit{dynamic recalibration}. 

Based on the above analysis, we propose a \textbf{Semantic Bridge Fusion Framework with Bi-Support Modeling} for multispectral object detection, as shown in Fig.~\ref{fig:teaser} (\textit{Semantic Bridge Fusion}). Specifically, by using text as a shared semantic bridge, we establish an RGB--IR response modeling mechanism under a unified category-semantic condition, disentangle the joint support into consensus support and discrepancy support, and perform structured modulation of IR semantic priors via dynamic recalibration, which then guides RGB fusion in a semantic-level manner. The \textit{Fusion Comparison} in Fig.~\ref{fig:teaser} shows that the joint support of real targets exhibits a clearer  target semantic distribution. In addition, considering that the IR modality is characterized by strong regional thermal responses but weak fine-grained textures, we design a dual-branch frequency-aware representation backbone to construct a thermal representation basis that is more suitable for subsequent fusion. We further build a bidirectional semantic alignment module, through which text guides visual enhancement and visual evidence in turn refines text, thereby forming a closed-loop vision--text collaboration. Meanwhile, the refined textual semantics are further used as semantic anchors for vision--text matching, enabling open-vocabulary detection. Fig.~\ref{fig:teaser} (\textit{Statistic Improvement}) shows that, compared with the baselines, our method achieves superior accuracy with substantially fewer parameters. Extensive experiments further demonstrate that our method attains state-of-the-art performance on multispectral benchmarks. The main contributions of this work are summarized as follows:

% Motivated by the above analysis, we propose a semantic-level semantic fusion framework for text-guided multispectral object detection. Specifically, by using text as a shared semantic bridge, we establish an RGB–IR response modeling mechanism under a unified category-semantic condition, disentangle the joint support into consensus support and discrepancy support, and perform structured modulation of IR semantic priors via dynamic recalibration, which then guides RGB fusion in a semantic-level manner. In addition, considering that the IR modality is characterized by strong regional thermal responses but weak fine-grained textures, we design a dual-branch frequency-aware representation backbone to construct a thermal representation basis that is more suitable for subsequent fusion. We further build a bidirectional semantic alignment module, through which text guides visual enhancement and visual evidence in turn refines text, thereby forming a closed-loop vision–text collaboration. Experimental results show that our method achieves new state-of-the-art performance on four multispectral datasets.

\begin{itemize}
   
 \item We propose a semantic-level multispectral fusion paradigm that uses text as a shared semantic bridge, upgrading cross-modal interaction from static aggregation to semantic-level support modeling under unified semantics.

 \item We introduce a structured inductive bias based on consensus support–discrepancy support, through which inconsistent support induced by modality differences is injected into the fusion process via a dynamic recalibration mechanism.

 % \item We construct a bidirectional semantic alignment module to achieve a closed-loop collaboration in which text guides visual enhancement and visual evidence refines text, and demonstrate state-of-the-art performance on three datasets.

 % \item We construct a bidirectional semantic alignment module to achieve a text-visual closed-loop refine collaboration; the refined textual semantics are further used as semantic anchors for vision--text matching to support open-vocabulary detection.

 % \item We construct a bidirectional semantic alignment module to establish a closed-loop vision--text collaboration, in which text guides visual enhancement and visual evidence refines textual semantics; the refined textual semantics are further used as semantic anchors for vision--text matching to support open-vocabulary detection.

 % \item We construct a bidirectional semantic alignment module to establish closed-loop vision--text collaboration for detection, and demonstrate state-of-the-art detection performance on three datasets.
 
 \item We construct a bidirectional semantic alignment module to establish closed-loop vision--text collaboration for multispectral detection, and demonstrate state-of-the-art performance on benchmark datasets.

 % the refined textual semantics are further used as semantic anchors for vision--category text matching to support open-vocabulary detection.

\end{itemize}

\section{Related Works}

\subsection{Multispectral Object Detection}

Multispectral object detection exploits complementary visible and infrared information for robust target perception.Early studies mainly focused on benchmark construction and basic dual-stream detection frameworks~\cite{Hwang2015KAIST,Liu2016HalfwayFusion}. More recently, text or high-level semantics have been introduced into multispectral detection beyond purely visual-correlation-based fusion~\cite{Zhang2020CFR,Zhang2021GAFF,Yun2022InfusionNet,He2023CALNet,Shen2024ICAFusion,Zeng2024MMIDet,Yuan2024CAGT}. Representative methods include cross-modal conflict-aware learning~\cite{He2023CALNet}, ICAFusion based on dual cross-attention Transformers~\cite{Shen2024ICAFusion}, and MMI-Det based on multimodal integration~\cite{Zeng2024MMIDet}.Recent studies have further explored the use of text or high-level semantic knowledge in multispectral detection. For example, Huang \emph{et al.} introduced semantic knowledge into few-shot multispectral detection~\cite{Huang2024FewShotSemantic}; Chen \emph{et al.} further adopted vision-text guidance to establish adaptive cross-modal fusion~\cite{Chen2025VLGuided}; and Li \emph{et al.} leveraged SAM-guided semantic knowledge to enhance semantic constraints in visible-infrared object detection~\cite{Li2025SAMGuided}. However, most existing methods still treat text as an auxiliary semantic enhancement signal or an external knowledge source, yielding limited effectiveness. In contrast, our method uses text as a shared semantic bridge and introduces consensus and discrepancy supports into fusion via dynamic recalibration for more effective multispectral detection.

\begin{figure*}[t!]
  \centering
  \includegraphics[width=0.8\textwidth,height=7cm]{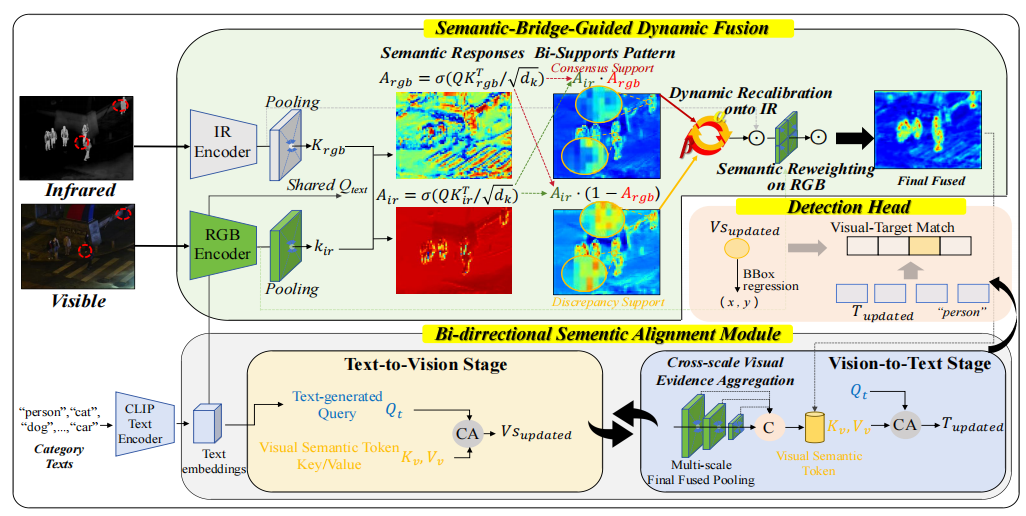}
  \vspace{-0.2cm}
      \caption{Overview of the proposed framework. It consists of a semantic-bridge-guided dynamic fusion module for modeling consensus and discrepancy support patterns for adaptive RGB--IR fusion, a bidirectional semantic alignment module for cross-modal semantic interaction, and a detection head for final prediction.}
  \Description{A woman and a girl in white dresses sit in an open car.}
  \label{fig:overview}
\end{figure*}

\subsection{Text-Visible-Infrared Fusion Paradigm}

Conventional text--vision fusion typically injects cross-modal information through semantic alignment, which has driven the development of recent vision foundation models and multimodal large models~\cite{Li2021ALBEF,Li2023BLIP2,Alayrac2022Flamingo}. In multispectral perception, text information has also been introduced into IR--RGB fusion~\cite{Liu2024SDCFusion,Zhou2024SeMIFusion,Sun2025CLIPFusion,Zhang2025TMCN}. Existing approaches mainly follow two directions. On the one hand, high-level task or scene semantics are fed back into fusion networks to improve adaptation to downstream tasks~\cite{Liu2024SDCFusion,Zhou2024SeMIFusion,Wang2024SDFuse,Yang2025KDFuse}, such as SeMIFusion~\cite{Zhou2024SeMIFusion}, SDFuse~\cite{Wang2024SDFuse}, and KDFuse~\cite{Yang2025KDFuse}. On the other hand, recent studies more directly introduce text prompts, vision--text models, or multimodal large models into fusion procedure~\cite{Qin2026TSPFusion,Yang2024MGFusion,Sun2025CLIPFusion,Zhang2025TMCN}. However, most existing methods still operate at the level of task adaptation or external enhancement, with limited attention to the positive role of textual semantics in guiding IR--RGB interaction. In contrast, our method highlights text as a shared semantic bridge to align IR and RGB and further guide their semantic mapping fusion.

\section{Problem Definition}

Text-guided multispectral object detection aims to exploit text semantics to organize cross-modal interaction between RGB and IR for more effective object detection. Due to the inherent differences between the two modalities in information granularity and scene adaptability, static RGB--IR aggregation at the pixel or feature level often fails to establish stable alignment under unified target semantics. Therefore, this work focuses on \textbf{semantic-level cross-modal support modeling under shared textual semantics}. Specifically, the text representation $T$ is no longer used merely for semantic enhancement or auxiliary prompting, but is treated as a \textbf{shared semantic anchor--bridge} that aligns RGB and IR around the same target semantics. Under the guidance of $T$, the two modalities produce text-conditioned semantic responses, denoted by $A_{rgb}$ and $A_{ir}$, respectively. Furthermore, considering that RGB and IR often exhibit missing single-modal evidence or inconsistent responses in complex scenes, we characterize semantic-level support from two complementary aspects: the \textbf{consensus support} $M_{cons}$, where the two modalities provide consistent evidence, and the \textbf{discrepancy support} $M_{dis}$, which arises from modality-specific differences. Accordingly, the core problem studied in this work is how to structurally model $A_{rgb}$ and $A_{ir}$ under a shared textual semantic anchor--bridge, and how to dynamically exploit $M_{cons}$ and $M_{dis}$ to obtain a more effective multispectral representation; meanwhile, text is further used as semantic anchors for vision-text matching to enable {open-vocabulary detection}.

\section{Methodology}

\subsection{Overview}
Fig. \ref{fig:overview} presents the overall pipeline of the proposed framework, which includes Semantic-Bridge-Guided Dynamic Fusion, Bidirectional Semantic Alignment, and Detection Head. The first module uses text as a shared semantic bridge to guide RGB–IR interaction and dynamically inject consensus and discrepancy supports into fusion, yielding semantically enhanced multispectral features. The second module further refines visual and textual semantics in a closed-loop manner to improve cross-modal consistency. The aligned vision-text features are then fed into the detection head, where text serves as semantic anchors for cross-modal matching, enabling open-vocabulary detection. In addition, we design an efficient infrared-specific backbone for thermal feature extraction, which is detailed in Section \ref{sec:OtherImportantComponents}.

\subsection{Semantic-bridge-Guided Dynamic Fusion for Multispectral Representation}

Existing RGB--IR detection methods mostly adopt pixel-level static fusion paradigms, which are not well aligned with the actual imaging characteristics of multispectral data. Specifically, the RGB modality is rich in fine-grained details, whereas the IR modality is more sensitive to target existence but provides relatively coarse representations. Given the inherent asymmetry between the two modalities in information granularity, a natural insight is that cross-modal fusion should not remain at uniform aggregation; instead, it should identify reliable target regions under a unified semantic reference and establish semantic-level cross-modal interaction accordingly. Based on this observation, we propose a semantic-anchor-guided semantic-level cross-modal recalibration mechanism, in which text serves as a shared semantic anchor, IR provides semantic-level spatial priors, and RGB carries fine-grained details, thereby enabling dynamically semantic-guided fusion toward target regions.

Specifically, given RGB features $X_{rgb}$, IR features $X_{ir}$, and text features $T$, we first map the text feature into a shared query $Q$, and map RGB and IR features into modality-specific keys:
\begin{equation}
Q = W_q T,\qquad
K_{rgb} = W_k^{rgb}(X_{rgb}),\qquad
K_{ir} = W_k^{ir}(X_{ir}).
\end{equation}

Then, under the unified textual semantic constraint, we compute the spatial responses of RGB and IR to the category semantics, respectively:
\begin{equation}
A_{rgb} = \sigma\!\left(\frac{QK_{rgb}^{\top}}{\sqrt{d_k}}\right),\qquad
A_{ir} = \sigma\!\left(\frac{QK_{ir}^{\top}}{\sqrt{d_k}}\right).
\end{equation}

Next, we establish the cross-modal consistent response relation via $A_{ir}A_{rgb}$, whose high-response regions correspond to the joint support of the two modalities for the same target semantics, and can therefore serve as a basis for target support identification.

\begin{figure}[b!]
  \centering
  \includegraphics[width=0.45\textwidth,height=4.5cm]{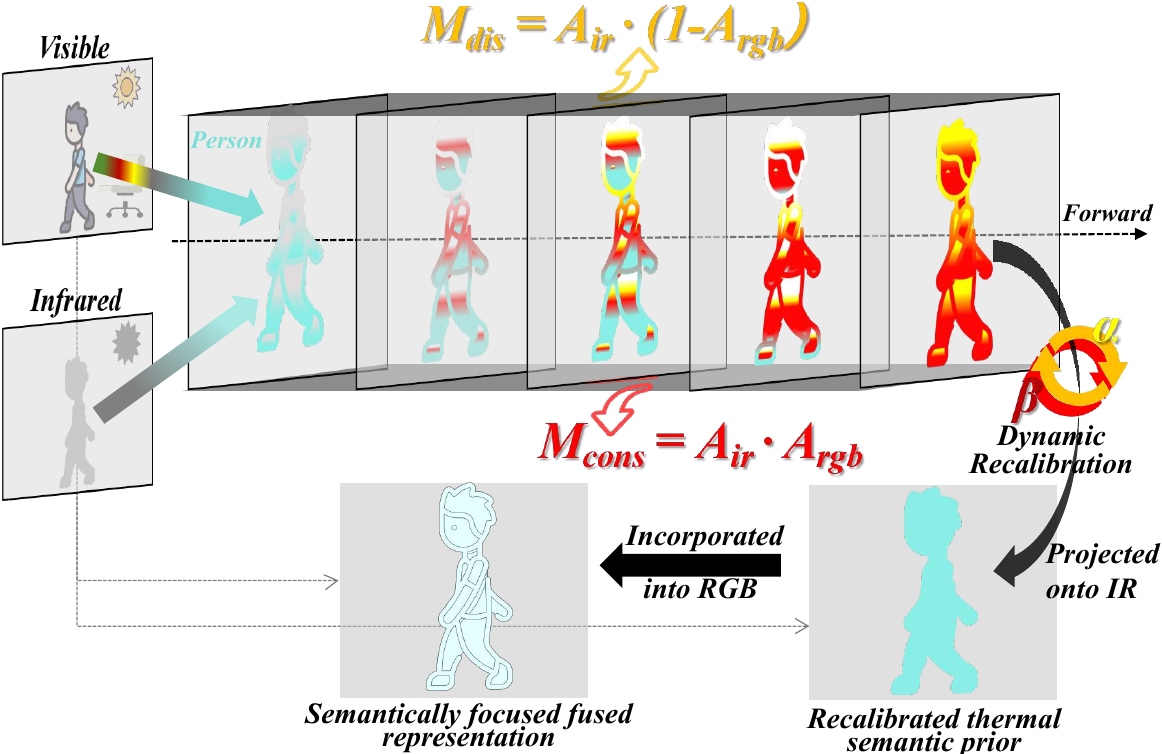}
  \caption{Illustration of the proposed bi-support modeling. Consensus and discrepancy support are constructed from semantic responses and jointly integrated through dynamic recalibration to enhance multispectral fusion.}
  \Description{A woman and a girl in white dresses sit in an open car.}
  \label{fig:bi-support modeling}
\end{figure}

However, as discussed in the problem definition, in addition to \emph{consensus}, \emph{discrepancy} caused by cross-modal inconsistency is also valuable to model. For example, under degraded conditions, IR may still preserve valid target cues, whereas RGB may fail to produce synchronized responses due to information loss, leading to semantically meaningful discrepancy support. Therefore, such regions should not be simply ignored, but rather used as an important cue for subsequent dynamic recalibration. To this end, we propose a consensus--discrepancy dynamic recalibration strategy, as shwon in \ref{fig:bi-support modeling}, to derive the consensus feature $M_{cons}$ for cross-modal consensus support and the discrepancy feature $M_{dis}$ for modal discrepancy support:
\begin{equation}
M_{cons} = A_{ir}A_{rgb},\qquad
M_{dis} = A_{ir}(1-A_{rgb}),
\label{eq:bi-support}
\end{equation}
\textit{s. t.}
\begin{equation}
A_{ir} = A_{ir}A_{rgb} + A_{ir}(1-A_{rgb}).
\end{equation}

Accordingly, the dynamic recalibration of IR features can be written as
\begin{equation}
\widetilde{X}_{ir}
=
X_{ir}\odot(1+\beta M_{cons})\odot(1+\alpha M_{dis}),
\end{equation}
where $\beta M_{cons}$ provides the dominant guidance from consensus support to reinforce stable target responses that are consistent across modalities, while $\alpha M_{dis}$ acts as a dynamic modulation term to supplement effective cues carried by discrepancy support and suppress potential discrepancy errors. Here, $\alpha$ and $\beta$ are learnable modulation coefficients optimized end-to-end to control the injection strength of the two types of support in feature recalibration. In particular, $\alpha$ functions as a gate on discrepancy support, so as to preserve potentially useful cues while suppressing discrepancy-induced errors.

After obtaining the recalibrated thermal semantic prior, we further transform it into dynamic guidance for RGB features. Specifically, we first align $\widetilde{X}_{ir}$ to the RGB channel space and generate a spatial attention map:
\begin{equation}
\widetilde{X}_{ir}^{align} = \psi(\widetilde{X}_{ir}),\qquad
W_{att} = \phi(\widetilde{X}_{ir}^{align}).
\end{equation}

Then, we use this attention map to perform semantic-level reweig\\hting on RGB features, and interact it with the aligned thermal prior via convolution:
\begin{equation}
\widehat{X}_{rgb} = X_{rgb}\odot W_{att},
\end{equation}
\begin{equation}
F_{fuse} = \operatorname{Conv}\big([\widehat{X}_{rgb},\,\widetilde{X}_{ir}^{align}]\big).
\end{equation}

In this way, we obtain a semantically focused semantic-level fused representation.

\subsection{Bidirectional Semantic Alignment Module}

To further enhance the consistency between multispectral visual features and category semantics for detection-head classification, we construct a bidirectional semantic alignment module. This module forms a closed-loop cross-modal interaction from two directions: \emph{text-to-vision} and \emph{vision-to-text}.

In the text-to-vision stage, since the fused visual features have already highlighted candidate features, the goal is no longer to assign category semantics to every spatial location, but to let each category retrieve its supporting evidence from the proposal-enhanced visual field. Accordingly, we adopt a cross-attention $\mathbf{A}_{t\to v}$ design with text-generated queries
\(
Q_t\in\mathbb{R}^{N\times d}
\)
and fused-feature-generated keys and values
\(
K_v\in\mathbb{R}^{HW\times d}
\)
and
\(
V_v\in\mathbb{R}^{HW\times C}
\).
If visual features are instead used as queries, the attention direction becomes $\mathbf{A}_{v\to t}$:
\begin{equation}
\mathbf{A}_{t\to v}\in\mathbb{R}^{N\times HW},\qquad
\mathbf{A}_{v\to t}\in\mathbb{R}^{HW\times N}.
\end{equation}
Here, \(\mathbf{A}_{t\to v}\) retrieves support regions over spatial positions for each category, which is more consistent with the current objective, namely, \emph{category semantics reading out candidate evidence}. By contrast, \(\mathbf{A}_{v\to t}\) is closer to position-wise semantic assignment, causing many low-information or background positions to participate in category modeling. 

% Additionally, textual semantics are injected into candidate regions in a spatially aware manner rather than as a single global vector, improving category discriminability while preserving the spatial structure required for detection.

In the vision-to-text stage, we further use visual evidence to update textual representations, so that text semantics evolve from static category-name embeddings into scene-conditioned dynamic semantic representations. Specifically, given multi-scale visual features
\(
\{P_i\}_{i=1}^{L}
\),
we construct a visual semantic token as
\begin{equation}
U=\operatorname{Concat}\big(\{\operatorname{Pool}(P_i)\}_{i=1}^{L}\big).
\end{equation}
Unlike text encoding based only on static category names, our text update is driven by multi-scale visual evidence from high-confidence target regions in the RGB-IR fused features rather than by background responses. As a result, the updated textual representation becomes more consistent with the current task distribution.

Through this closed-loop process of \emph{text-guided visual enhancement} and \emph{vision-guided text refinement}, vision and text achieve bidirectional and adaptive collaborative alignment in a shared semantic space, which further benefits downstream detection-head classification.

\subsection{Other Important Components}
\label{sec:OtherImportantComponents}
In addition to the two core modules, we further introduce several auxiliary components for efficient infrared representation and detection.

\subsubsection{Detection Head}
After bidirectional semantic alignment, the resulting vision-text features are fed into a detection head. Specifically, visual features are used for region localization and box regression, while object classification is performed by matching region representations with the updated textual embeddings. Let \(R=\{r_i\}_{i=1}^{N_r}\) denote the visual region features and let \(\widetilde{T}=\{\tilde{t}_j\}_{j=1}^{N_c}\) denote the updated text embeddings:
\(
z_i^{v}=\mathcal{P}_{v}(r_i),\quad
z_j^{t}=\mathcal{P}_{t}(\tilde{t}_j).
\)
Their semantic similarity is computed by
\begin{equation}
s_{ij}=\frac{{z_i^{v}}^{\top} z_j^{t}}{\|z_i^{v}\|\,\|z_j^{t}\|}.
\end{equation}
We then optimize the region--text matching with a contrastive objective:
\begin{equation}
\mathcal{L}_{\text{match}}
=
-\frac{1}{|\mathcal{P}|}
\sum_{(i,j)\in\mathcal{P}}
\log
\frac{\exp(s_{ij}/\tau)}
{\sum_{k=1}^{N_c}\exp(s_{ik}/\tau)},
\end{equation}
where \(\mathcal{P}\) denotes the set of matched region--text pairs and \(\tau\) is a temperature parameter. In this way, object categories are inferred through similarity to textual semantics, which also allows the detector to generalize beyond a fixed closed-set label space.

\subsubsection{Dual-Branch Frequency-Aware IR Backbone}
As an auxiliary yet important component, we design a lightweight frequency-aware backbone for infrared representation. Unlike directly reusing the visible-light encoder or adopting a single-branch design, this backbone explicitly separates shallow IR features into low-frequency thermal distribution and high-frequency contour cues, so as to preserve stable heat responses while compensating for weak local texture representation.

Given the shallow IR feature
\(
X_{ir}^{0}\in\mathbb{R}^{C\times H\times W},
\)
we first transform it into the frequency domain and decompose it with a learnable mask:
\begin{equation}
\hat{X}_{ir}^{0}=\mathcal{D}(X_{ir}^{0}),\qquad
X_{ir}^{L}=\mathcal{D}^{-1}(M\odot \hat{X}_{ir}^{0}),\quad
X_{ir}^{H}=\mathcal{D}^{-1}((1-M)\odot \hat{X}_{ir}^{0}),
\end{equation}
where \(M\in[0,1]^{H\times W}\) is a learnable frequency mask. For efficiency, this decomposition is only applied to shallow IR features.

The two components are then encoded and recombined as
\begin{equation}
\widetilde{X}_{ir}^{0}=\phi_L(X_{ir}^{L})+\phi_H(X_{ir}^{H}),
\end{equation}
where the low-frequency branch captures regional thermal distribution, while the high-frequency branch enhances local contours and structures. We further adopt SE operations for efficient channel recalibration.

\begin{table*}[t!]
\centering
\small
\setlength{\tabcolsep}{6pt}
\renewcommand{\arraystretch}{1.08}
\caption{Comparison with state-of-the-art methods on the LLVIP and FLIR datasets. Results are reported in mAP, mAP@0.5, and mAP@0.75. AVG denotes the average of the overall accuracy on LLVIP and FLIR. Best results are in bold. }
\vspace{-0.2cm}
\label{tab:llvip_flir_comparison}
\resizebox{\textwidth}{!}{
\begin{tabular}{l c c c ccc ccc c c}
\toprule
\multirow{2}{*}{Method} & \multirow{2}{*}{Modality} & \multirow{2}{*}{Year} & \multirow{2}{*}{Backbone}
& \multicolumn{3}{c}{LLVIP} & \multicolumn{3}{c}{FLIR} & \multirow{2}{*}{AVG$\uparrow$} & \multirow{2}{*}{Params/M$\downarrow$} \\
\cmidrule(lr){5-7} \cmidrule(lr){8-10}
&&&& mAP$\uparrow$ & mAP@0.5$\uparrow$ & mAP@0.75$\uparrow$ & mAP$\uparrow$ & mAP@0.5$\uparrow$ & mAP@0.75$\uparrow$ && \\
\midrule
FasterRCNN\cite{NIPS2015_14bfa6bb}         & RGB    & 2015 & ResNet50           & 45.1  & 87.0 & 41.2  & 27.7 & 62.2 & 21.2 & 36.4 & 41.1 \\
RetinaNet\cite{RetinaNet}          & RGB    & 2017 & ResNet50           & 42.8  & 88.0 & 34.4  & 21.9 & 51.2 & 15.2 & 32.4 & 43.0 \\
YOLOv5 \cite{yolov5}          & RGB    & 2020 & CSPDarknet       & 54.0  & 91.9 & 52.5  & 28.2 & 66.3 & 24.2 & 41.1 & 47.6 \\
DDQ-DETR \cite{DDQ-DETR}        & RGB    & 2023 & ResNet50           & 46.7  & 86.1 & 45.8  & 30.9 & 64.9 & 24.5 & 38.8 & 244.6 \\
\midrule
FasterRCNN \cite{NIPS2015_14bfa6bb}        & IR     & 2015 & ResNet50           & 54.5  & 94.6 & 57.6  & 37.6 & 75.8 & 31.6 & 46.1 & 41.1 \\
RetinaNet \cite{RetinaNet}         & IR     & 2017 & ResNet50         & 55.1  & 94.8 & 57.6  & 31.5 & 66.1 & 25.3 & 43.3 & 43.0 \\
YOLOv5 \cite{yolov5}            & IR     & 2020 & CSPDarknet       & 62.1  & 95.2 & 67.0  & 38.3 & 72.9 & 31.8 & 50.2 & 47.6 \\
DDQ-DETR  \cite{DDQ-DETR}         & IR     & 2023 & ResNet50           & 58.6  & 93.9 & 64.6  & 37.1 & 73.9 & 32.2 & 47.9 & 244.6 \\
\midrule
ICAFusion\cite{Shen2024ICAFusion}      & RGB+IR & 2024 & CSPDarknet       & --    & --   & --    & 41.4 & 79.2 & 36.9 & --   & 120.0 \\
RSDet \cite{RSDet}        & RGB+IR & 2024 & ResNet50           & 61.3  & 95.8 & 70.4  & 43.8 & 83.9 & 40.1 & 52.6 & -- \\
% UniRGB-IR [16]     & RGB+IR & 2025 & ViT              & 63.2  & 96.1 & 72.2  & 44.1 & 81.4 & 40.2 & 53.7 &\textbf{ 8.9} \\
CrossModalNet \cite{CrossModalNet} & RGB+IR & 2025 & CSPDarknet       & 64.7  & 97.7 & 73.5  & 43.3 & 81.7 & 39.1 & 54.0 & 92.8 \\
COFNet \cite{COFNet}        & RGB+IR & 2025 & CSPDarknet       & 65.9  & 97.7 & 75.9  & 44.6 & 83.6 & 41.7 & 55.3 & 90.2 \\
Fusion Mamba \cite{FusionMamba}      & RGB+IR & 2025 & CSPDarknet       & 64.3  & 97.0 & --    & 47.0 & 84.9 & \textbf{45.9} & 55.7 & 287.6 \\
waveMamba \cite{WaveMamba}        & RGB+IR & 2025 & CSPDarknet       & 66.0  & \textbf{98.3} & --    & 48.1 & 88.4 & --   & 57.1 & 69.1 \\
MCOR \cite{MCOR}              & RGB+IR & 2025 & CSPDarknet       & 64.9  & 97.6 & 73.6  & 39.9 & 78.2 & 34.4 & 52.4 & -- \\
SLGNet  \cite{SLGNets}           & RGB+IR & 2026 & ViT-based        & 66.1  & \textbf{98.3} & 75.4  & 45.1 & 85.8 & 42.3 & 55.6 & \textbf{12.1} \\
MS2Fusion \cite{Ms2fusion}      & RGB+IR & 2026 &  CSPDarknet  & 65.5  & 97.5 & -  & 40.3 & 83.3 & - &  52.9 & 130.2 \\
MDACL \cite{MDACL}           & RGB+IR & 2026 & CSPDarknet       & 66.5  & 97.9 & --    & 44.6 & 83.2 & --   & 55.6 & -- \\
\rowcolor{gray!30} {Ours}      & {RGB+IR} & {2026} & {CSPDarknet} & \textbf{68.9} & \textbf{98.3} & \textbf{81.0} & \textbf{50.4} & \textbf{88.6} & 45.7 & \textbf{59.7} & {15.5} \\
\bottomrule
\end{tabular}
}
\end{table*}

\begin{table*}[t!]  
\centering
\small
\setlength{\tabcolsep}{12pt}
\renewcommand{\arraystretch}{1.08}
\caption{Comparison with state-of-the-art methods on the DroneVehicle dataset. Results are reported for each category together with mAP@0.5. Best results are in bold.}
\vspace{-0.2cm}
\label{tab:dronevehicle_comparison}
\resizebox{\textwidth}{!}{
\begin{tabular}{l c c c c c c c c c}
\toprule
\multirow{2}{*}{Method} & \multirow{2}{*}{Modality} & \multirow{2}{*}{Year} & \multirow{2}{*}{Backbone}
& \multicolumn{5}{c}{Per-class AP} & \multirow{2}{*}{mAP@0.5$\uparrow$} \\
\cmidrule(lr){5-9}
&&&& Car$\uparrow$ & Freight-Car$\uparrow$ & Truck$\uparrow$ & Bus$\uparrow$ & Van$\uparrow$ & \\
\midrule

RetinaNet \cite{RetinaNet}  & RGB    & 2017 & ResNet50   & 80.2 & 16.1 & 29.0 & 53.5 & 18.4 & 39.4 \\
Faster R-CNN\cite{NIPS2015_14bfa6bb} & RGB    & 2015 & ResNet50   & 88.8 & 40.9 & 57.2 & 78.6 & 42.0 & 61.7 \\
ReDet \cite{ReDet}       & RGB    & 2021 & ResNet50   & 80.3 & 42.7 & 56.1 & 80.2 & 44.4 & 60.8 \\
\midrule

RetinaNet\cite{RetinaNet}    & IR     & 2017 & ResNet50   & 89.3 & 40.0 & 38.2 & 79.0 & 32.1 & 55.7 \\
Faster R-CNN\cite{NIPS2015_14bfa6bb} & IR     & 2015 & ResNet50   & 89.4 & 48.3 & 53.5 & 87.0 & 42.6 & 64.2 \\
ReDet   \cite{ReDet}     & IR     & 2021 & ResNet50   & 89.8 & 43.3 & 53.9 & 84.8 & 33.2 & 61.0 \\
\midrule

SLBAF-Net \cite{SLBAF-Net}   & RGB+IR & 2023 & CSPDarknet & 90.2 & 72.0 & 68.6 & 89.9 & 59.9 & 76.1 \\
C$^2$Former \cite{C2Former} & RGB+IR & 2024 & ResNet50   & 90.2 & 68.3 & 64.4 & 89.8 & 58.5 & 74.2 \\
M2FP \cite{M2FP}        & RGB+IR & 2024 & ViT        & 95.7 & 64.7 & 76.2 & 92.1 & 64.7 & 78.4 \\
GLFNet \cite{GLFNet}      & RGB+IR & 2024 & ResNet50   & 90.3 & 53.6 & 72.7 & 88.0 & 52.6 & 71.4 \\
YOLOFIV \cite{YoloFiv}     & RGB+IR & 2024 & CSPDarknet & 95.9 & 34.6 & 64.2 & 91.6 & 37.5 & 64.7 \\
AFFNet  \cite{AFFNet}    & RGB+IR & 2025 & ResNet50   & 90.2 & 56.0 & 59.5 & 89.0 & 53.7 & 69.7 \\
% MGFF         & RGB+IR & 2025 & CSPNeXt    & 90.4 & 69.6 & 80.9 & 89.9 & 68.0 & 79.8 \\
DMM  \cite{DMM}        & RGB+IR & 2025 & ResNet50   & 90.4 & 63.0 & \textbf{77.8} & 88.7 & \textbf{66.0} & 77.2 \\
% WaveMamba    & RGB+IR & 2025 & CSPDarknet & 95.0 & 68.5 & 80.4 & 90.6 & 64.5 & 79.8 \\
CRSIOD\cite{CRSIOD}       & RGB+IR & 2025 & CSPDarknet & 94.5 & 46.3 & 63.7 & 92.0 & 45.7 & 68.4 \\
% SLGNet       & RGB+IR & 2026 & ViT        & 96.1 & 69.4 & 80.9 & 91.8 & 65.3 & 80.7 \\
\rowcolor{gray!30}{Ours} & {RGB-IR} & {2026} & {CSPDarknet}
             & \textbf{96.9} & \textbf{71.1} & 77.7 & \textbf{95.8} & 65.2 & \textbf{81.4} \\
\bottomrule
\end{tabular}
}
\end{table*}

\section{Experiments}

\subsection{Experimental Setup}

\paragraph{Datasets.}
We evaluate the proposed method on four public multispectral detection benchmarks, including FLIR\cite{flir_adas_dataset}, LLVIP\cite{jia2021llvip}, DroneVehicle\cite{sun2022dronevehicle}, and M3FD\cite{liu2022target}. FLIR is a widely used RGB--thermal benchmark for autonomous driving scenarios. LLVIP is a paired visible--infrared dataset designed for low-light vision and pedestrian perception. DroneVehicle is a UAV-based RGB--infrared benchmark for vehicle detection under aerial viewpoints. M3FD is a multi-scenario multi-modality benchmark that provides aligned visible--infrared image pairs for object detection. For evaluation, we report mAP, including mAP at IoU thresholds of 0.5 and 0.75.

\paragraph{Implementation Details.}
Our method is implemented in \texttt{PyTorch} and trained on a single \texttt{NVIDIA H800 GPU} with 80\,GB memory. For the RGB branch, we adopt a YOLO-series \texttt{CSPDarknet} backbone to extract hierarchical visible features. We use \texttt{AdamW} as the optimizer, with an initial learning rate of $1\times10^{-4}$. 

% The code of our method has been released at \url{https://github.com/zhenwang5372/Bridging-RGB-IR-Gap}.

\begin{figure*}[t!]
  \centering
  \includegraphics[width=\textwidth,height=9.5cm]{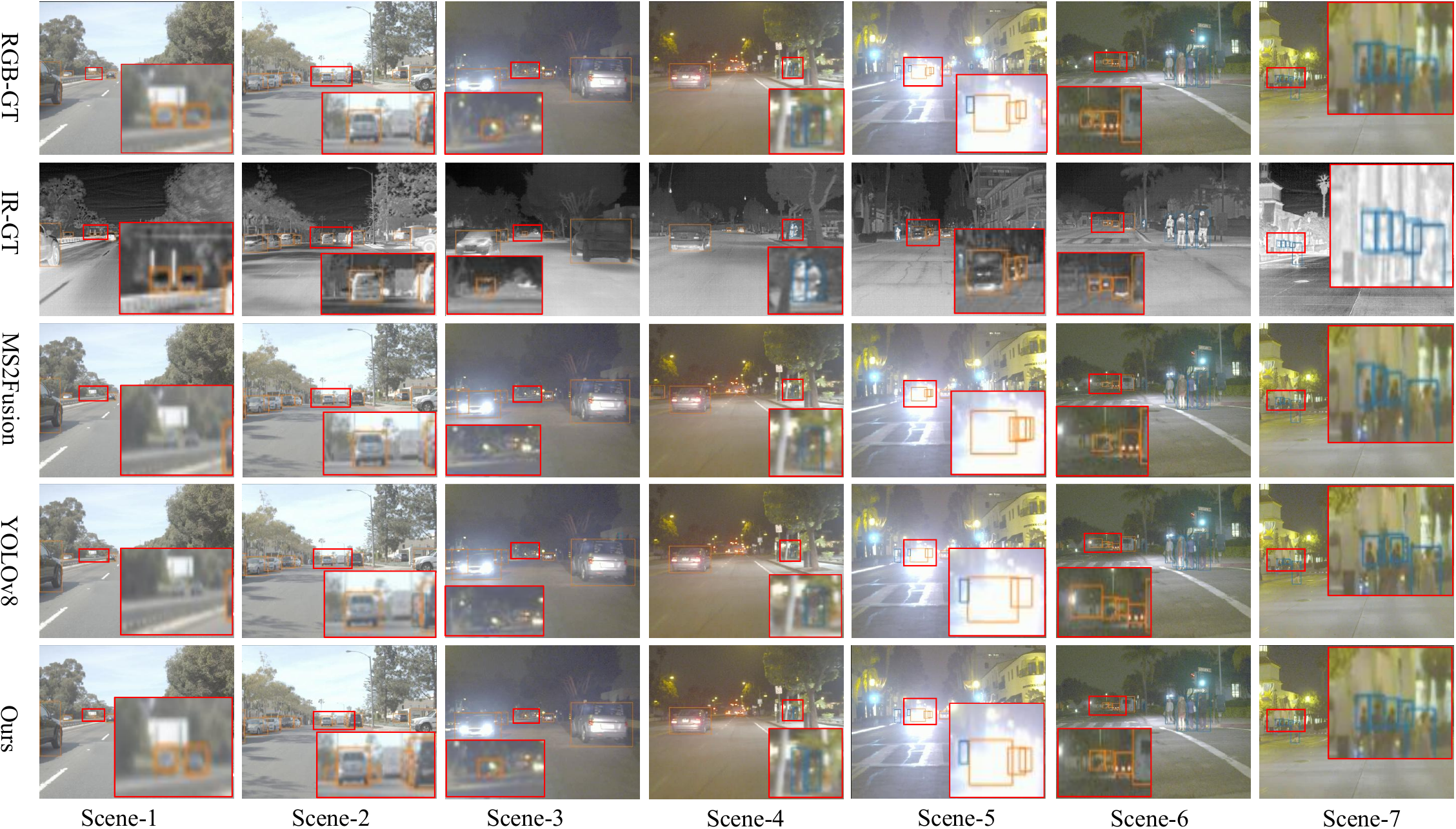}
  \vspace{-0.7cm}
  \caption{Qualitative comparison on seven representative scenes from the FLIR dataset. The examples include daytime perception (Scene-1\&2), small target perception in nighttime conditions (Scene-3), occluded cases (Scenes-4\&5), and crowded scenes (Scenes-6\&7). Our method yields more reliable predictions in challenging multispectral scenes.}
  \Description{A woman and a girl in white dresses sit in an open car.}
  \label{fig:visual-filr}
\end{figure*}

\paragraph{Baselines.}
To thoroughly evaluate the proposed method, we conduct comprehensive comparisons with state-of-the-art object detection models. The compared methods include both single-modality baselines (IR-only or RGB-only) and multi-modality approaches (IR+RGB), enabling a systematic evaluation under different input settings. The results reported are directly taken from the original papers whenever available, and are compared under consistent experimental settings to ensure fairness and objectivity. More detailed experimental setups are provided in Appendix A.

\subsection{Comparative Experiments}

% \paragraph{LLVIP \& FLIR} Table~\ref{tab:llvip_flir_comparison} reports the comparison results on the LLVIP and FLIR datasets. Overall, the proposed method achieves consistently strong performance across both benchmarks and attains the best cross-dataset average mAP of 59.7. On LLVIP, compared with the strongest previous multi-modality method, CAFF\_DI\\NO, our method further improves the overall mAP from 68.5 to 68.95 (+0.45), while achieving a larger gain on mAP@0.75, from 79.0 to 80.95 (+1.95, 2.47\% relatively). On FLIR, although CAFF\_DINO reports a slightly higher overall mAP of 50.5, our method achieves the best mAP@0.5 among all compared methods. More importantly, among CSPDarknet-based multi-modality methods, our method surpasses the strongest prior result on FLIR (waveMamba, 48.1 mAP) by 2.3 points, yielding a relative improvement of 4.78\%. Moreover, despite using only 15.5M parameters, our method remains highly competitive against substantially larger models (Fusion Mamba, 287.6M), demonstrating a favorable trade-off between detection performance and model complexity. These results verify that the proposed framework provides a favorable balance between effectiveness and generalization across different multispectral benchmarks.

\paragraph{LLVIP \& FLIR}
On the LLVIP and FLIR benchmarks (Table \ref{tab:llvip_flir_comparison}), our method achieves the best accuracy performances while maintaining a substantially reduced parameter count. For example, compared with MS2Fusion (2026), ours delivers a 12.9\% relative gain in average mAP together with an 88.1\% reduction in parameter count. In addition, despite having only a slightly larger Params than SLGNet (2026), ours achieves an 11.8\% relative improvement in FLIR mAP. Fig.~\ref{fig:visual-filr} shows representative visual  examples from the FLIR dataset. Compared with the baselines, our method delivers more reliable perception, especially for distant small objects, across daytime scenes (Scenes-1\&2), nighttime conditions (Scene-3), occluded cases (Scenes-4\&5), and crowded scenes (Scenes-6\&7). It also detects unlabeled small pedestrians missed by the ground truth (Scene-1), suggesting stronger generalization ability.

\paragraph{DroneVehicle.}
The superior performance of our method on the DroneVehicle (Table~\ref{tab:dronevehicle_comparison}) further verifies its scene generalization ability. Under the same CSPDarkNet backbone, our method improves the overall mAP@0.5 over CRSIOD by 19.0\%, and also surpasses the overall second-best method M2FP by 3.8\%, further demonstrating its effectiveness for drone-view small-object detection.

\paragraph{M3FD}
Table~\ref{tab:backbone_comparison} shows that our method consistently achieves the best accuracy on M3FD. Notably, even when replacing the RGB backbone, our method still maintains superior performance, demonstrating strong generalizability across both datasets and backbone architectures. More comparative experimental results and detailed analyses are provided in Appendix B.

\begin{table}[t]
\centering
\small
\setlength{\tabcolsep}{4pt}
\renewcommand{\arraystretch}{1.08}
\vspace{-0.3cm}
\caption{Comparison of different methods under ResNet50 and CSPDarknet backbones on the M3FD dataset.}
\vspace{-0.3cm}
\label{tab:backbone_comparison}
\resizebox{\columnwidth}{!}{
\begin{tabular}{l l c c c c c c}
\toprule
Backbone & Method & Modality & Car$\uparrow$ & Bus$\uparrow$ & Truck$\uparrow$ & mAP$_{50}$$\uparrow$ & mAP$\uparrow$ \\
\midrule
\multirow{6}{*}{ResNet50} 
& RetinaNet \cite{RetinaNet}     & IR      & 79.0 & 68.0 & 60.0 & 57.9 & 33.8 \\
& Faster-RCNN\cite{NIPS2015_14bfa6bb}    & IR      & 81.3 & 66.7 & 63.3 & 63.3 & 35.6 \\
& RetinaNet \cite{RetinaNet}     & RGB     & 83.2 & 75.8 & 69.9 & 64.7 & 37.4 \\
& Faster-RCNN \cite{NIPS2015_14bfa6bb}   & RGB     & 87.9 & 81.3 & 72.4 & 74.5 & 42.3 \\
& TiNet  \cite{TiNet}        & RGB+IR  & 87.5 & 83.1 & 81.6 & 79.8 & 49.5 \\
% &\rowcolor{gray!30}  {Ours}  & RGB+IR  & \textbf{91.0} & \textbf{92.7} & \textbf{87.7} & \textbf{84.3} & \textbf{55.8} \\
% ResNet50 的 Ours 行
& \cellcolor{gray!30} {Ours} & \cellcolor{gray!30} RGB+IR & \cellcolor{gray!30} \textbf{91.0} & \cellcolor{gray!30} \textbf{92.7} & \cellcolor{gray!30} \textbf{87.7} & \cellcolor{gray!30} \textbf{84.3} & \cellcolor{gray!30} \textbf{55.8} \\
\midrule
\multirow{6}{*}{CSPDarknet} 
& YPLOV5l \cite{yolov5}       & IR      & 87.2 & 86.8 & 82.5 & 80.3 & 50.7 \\
& YOLOv8l   \cite{yolov8_ultralytics}     & IR      & 90.0 & 90.9 & 85.9 & 79.5 & 53.1 \\
& YPLOV5l  \cite{yolov5}      & RGB     & 90.5 & 91.2 & 86.1 & 82.7 & 52.5 \\
& YOLOv8l   \cite{yolov8_ultralytics}     & RGB     & 91.2 & 92.9 & 86.0 & 80.9 & 52.5 \\
& RI-YOLO  \cite{RI-YOLO}     & RGB+IR  & 83.6 & 56.6 & \textbf{91.6} & 88.5 & \textbf{87.8} \\
% & \rowcolor{gray!30} {Ours}  & RGB+IR  & \textbf{95.0} & \textbf{94.6} & 89.7 & \textbf{91.6} & 63.5 \\
& \cellcolor{gray!30} {Ours}  & \cellcolor{gray!30} RGB+IR  & \cellcolor{gray!30} \textbf{95.0} & \cellcolor{gray!30} \textbf{94.6} & \cellcolor{gray!30} 89.7 & \cellcolor{gray!30} \textbf{91.6} & \cellcolor{gray!30} 63.5 \\
\bottomrule
\end{tabular}
}
\end{table}

\subsection{Ablation Studies}
The overall ablation study (Table \ref{tab:overall_ablation}) validates the efficacy of the proposed architecture, demonstrating that the core Semantic-bridge-Guided Dynamic Fusion module yields a remarkable mAP improvement over the V0 baseline. 

\begin{table}[h!]
\centering
\small
\setlength{\tabcolsep}{2pt}
\renewcommand{\arraystretch}{1.08}
\vspace{-0.3cm}
\caption{Overall ablation study of the proposed framework on the FLIR dataset. Best results are in bold.}
\vspace{-0.3cm}
\label{tab:overall_ablation}
\resizebox{\columnwidth}{!}{
\begin{tabular}{l c c c c c c}
\toprule
Variant & IR Backbone & Dynamic Fusion & Bidirectional Alignment & recall$\uparrow$ & mAP$_{50}$ $\uparrow$ & mAP$\uparrow$ \\
\midrule
V0  &            &                &                & 0.6845 & 0.775 & 0.383 \\
V1          & \checkmark &                &                & 0.6982 & 0.789 & 0.398 \\
V2          & \checkmark & \checkmark     &                & 0.7663 & 0.871 & 0.488 \\
V3 (Full)   & \checkmark & \checkmark     & \checkmark     & \textbf{0.7785} & \textbf{0.886} & \textbf{0.504} \\
\bottomrule
\end{tabular}
}
\end{table}

Further in-depth analytical experiments regarding specific functionalities are presented below, with the detailed evaluation of the IR backbone relegated to the Appendix C.

\subsection{Analytical Experiments}
We further conduct a series of analytical experiments to validate the effectiveness of the core components and key functionalities of the proposed framework.

\begin{table}[h!]
\centering
\small
\setlength{\tabcolsep}{6pt}
\renewcommand{\arraystretch}{1.08}
\caption{Analysis of the two roles of text on the FLIR dataset. Best results are in bold.}
\label{tab:text_role_analysis}
\resizebox{0.95\columnwidth}{!}{
\begin{tabular}{l c c c c}
\toprule
Method Variant & AP$\uparrow$ & AP$_{50}$$\uparrow$ & AP$_{75}$$\uparrow$ & FLOPs$\downarrow$ \\
\midrule
\multicolumn{5}{c}{\textit{Text as a Semantic Bridge}} \\
Vanilla Direct RGB--IR Fusion     & 41.9 & 82.2 & 36.4 & 13.56 G    \\
Conditional Prompt Fusion & 41.5 & 83.4 & 31.7 & 13.56 G    \\
Ours (Semantic Bridge Fusion)   & \textbf{50.6} & \textbf{88.6} & \textbf{45.7} & \textbf{4.05G}   \\
\midrule
\multicolumn{5}{c}{\textit{Text as a Semantic Anchor}} \\
w/o Bidirectional Update        & 46.4 & 85.4 & 41.6 & --   \\
Fixed Random Text Embedding   & 42.6 & 82.6 & 38.9 & --   \\
Ours                            & \textbf{50.6} & \textbf{88.6} & \textbf{45.7} & --   \\
\bottomrule
\end{tabular}
}
\end{table}

\textbf{{Semantic Guidance of Text.}} We further validate the role of \textit{Text} in our framework. 

\begin{figure}[h!]
  \centering
  \includegraphics[width=0.48\textwidth,height=1.9cm]{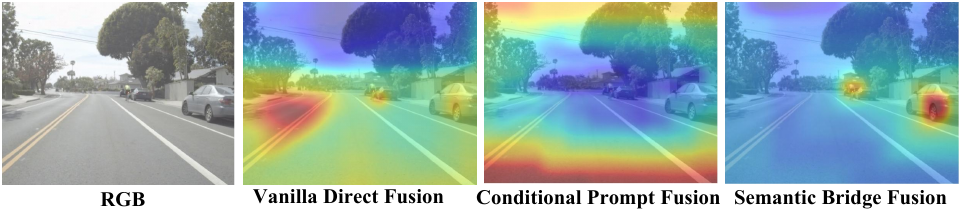}  
  \vspace{-0.4cm}
  \caption{Perception performance comparison on FLIR. The proposed fusion achieves better target perception than vanilla direct fusion and conditional prompt fusion.}
  \label{fig:fusion-comparison}
\end{figure}

\paragraph{Semantic Bridge Role of Text.} We validate the text's role as a semantic bridge against two classic multispectral fusion paradigms. As visualized in Fig.~\ref{fig:fusion-comparison}, our method exhibits superior target-perception capabilities. Quantitatively (Table \ref{tab:text_role_analysis}), it achieves the highest detection accuracy. Crucially, the semantic-bridge design acts as an information bottleneck that circumvents the high complexity of direct RGB-IR computations, reducing FLOPs by 70.1\%.

\vspace{-0.2cm}

\paragraph{Semantic Anchor Role of Text.}
We further validate the role of text as a semantic anchor in bidirectional semantic alignment. To this end, we consider two variants: removing the bidirectional update and replacing the updated text embedding with a fixed random embedding. Results in Table \ref{tab:text_role_analysis} in terms of accuracy consistently show inferior performance, verifying the importance of text anchoring and dynamic semantic update for detection.

\begin{table}[h!]
\centering
\small
\setlength{\tabcolsep}{1pt}
\renewcommand{\arraystretch}{1.08}
\caption{Effectiveness of the bi-support modeling on FLIR. ``Consensus'' and ``Discrepancy'' denote the two support patterns, respectively. ``Recalibration'' denotes dynamic recalibration. Best results are in bold.}
\vspace{-0.3cm}
\label{tab:bi_support_ablation}
\resizebox{0.85\columnwidth}{!}{
\begin{tabular}{c c c c c c}
\toprule
Consensus Support & Discrepancy Support & Recalibration & mAP$\uparrow$ & mAP$_{50}$$\uparrow$ & mAP$_{75}$$\uparrow$ \\
\midrule
 &          &          & 41.3 & 80.2 & 36.1 \\
\checkmark &          &          & 42.2 & 82.5 & 37.4 \\
          & \checkmark &          & 44.5 & 82.5 & 38.9 \\
\checkmark & \checkmark &          & 42.6 & 80.5 & 37.6 \\
\checkmark & \checkmark & \checkmark & \textbf{50.4} & \textbf{88.6} & \textbf{45.7} \\
\bottomrule
\end{tabular}
}
\end{table}

\textbf{Effectiveness of Bi-Support Modeling.} We further analyze the effectiveness of the proposed bi-support modeling from both component-wise and degradation-aware perspectives.

\paragraph{Component-wise Analysis.}

We first investigate the effectiveness of the proposed bi-support modeling on FLIR (Table~\ref{tab:bi_support_ablation}). Results demonstrate that both {consensus support} $M_{cons}$ and {discrepancy support} $M_{dis}$ patterns are individually effective. Furthermore, dynamic recalibration adaptively adjusts the application of the bi-support, which effectively further enhances bi-support complementarity, ultimately boosting the mAP to 50.4\%.

\begin{figure}[h!]
  \centering
  \includegraphics[width=0.25\textwidth,height=5cm]{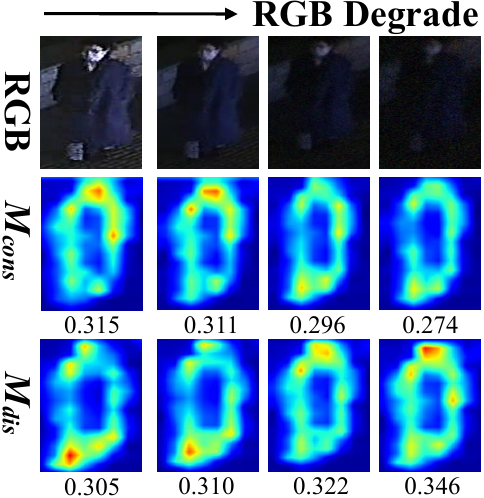}
  \vspace{-0.3cm}
  \caption{Visualization of responses of {consensus support} $M_{cons}$ and {discrepancy support} $M_{dis}$ under progressively degraded RGB inputs on a typical sample. The numbers denote average activation values within the ground-truth box.}
  \Description{A woman and a girl in white dresses sit in an open car.}
  \label{fig:degrade}
\end{figure}

\begin{figure}[h!]
  \centering
  \includegraphics[width=0.45\textwidth,height=4cm]{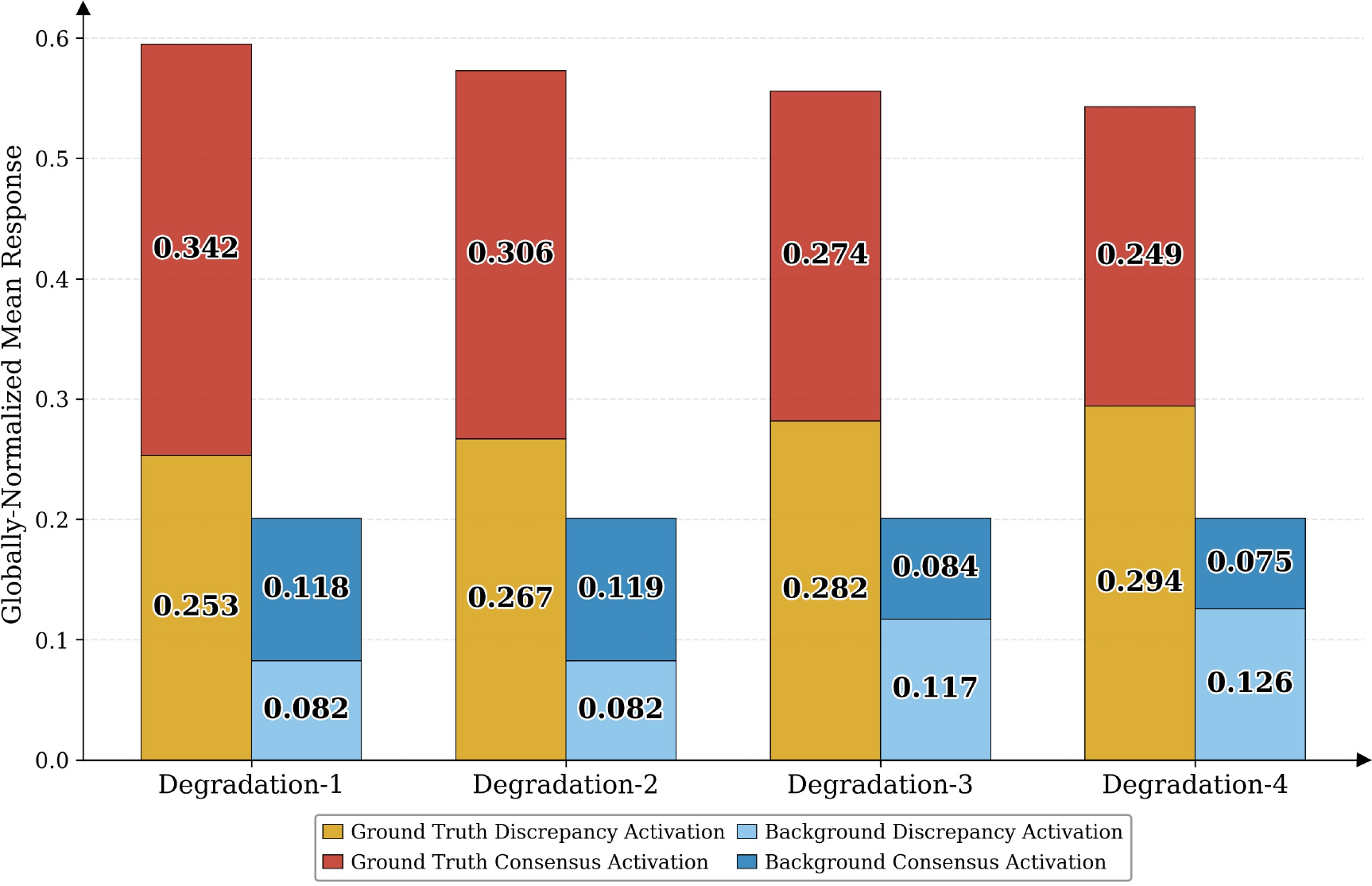}  
  \vspace{-0.3cm}
  \caption{Population-level trends of the two support patterns across degradation levels. The numbers denote the average activation values within the GT-box or Background.}
  \Description{A woman and a girl in white dresses sit in an open car.}
  \label{fig:degrade-200}
\end{figure}

\paragraph{Behavior under Progressive Degradation.}

To further validate the necessity and complementarity of the bi-support modeling, we examine its behavior under progressively degraded scenarios from both instance-level and population-level perspectives on the LLVIP dataset. As shown in Fig.~\ref{fig:degrade}, the activation values of $M_{\text{cons}}$ inside the ground-truth box gradually decrease as RGB quality deteriorates, whereas those of $M_{\text{dis}}$ increase. This trend indicates that the contributions of the two support patterns vary with degradation severity and exhibit clear complementarity. Specifically, consensus support dominates when cross-modal commonality is reliable, while discrepancy support becomes more important when degradation weakens the RGB cues. Fig.~\ref{fig:degrade-200} further illustrates, from population-level perspective, that as degradation increases, the response of $M_{\text{cons}}$ in GT-box gradually decreases, while that of $M_{\text{dis}}$ correspondingly increases, showing their complementary roles. These results validate the necessity of modeling both support patterns. Considering the degraded scenarios typical of multispectral detection, our design holds significant promise for practical deployment. Detailed degradation experiment settings and  analysis in the Appendix C.

\section{Conclusion}
In this paper, we present a novel text-guided multispectral object detection framework that rethinks RGB--IR interaction from a semantic perspective. Instead of treating text as a conditional prompt, we exploit it as a shared semantic bridge to guide cross-modal fusion, enabling more structured interaction between visible and thermal modalities. Building on this design, we further introduce bi-support pattern modeling to explicitly characterize both consensus and discrepancy cues across modalities, together with dynamic recalibration for adaptive fusion. In addition, a bidirectional semantic alignment strategy is developed to establish tighter collaboration between visual features and textual semantics, while a lightweight frequency-aware infrared backbone is used to better characterize thermal patterns. Extensive experiments on multispectral benchmarks demonstrate the effectiveness and generalization ability of the proposed framework.

% To improve cross-scale thermal semantics, we further perform lightweight top-down aggregation over multi-scale features \(\{F_i^{ir}\}_{i=1}^{L}\):
% \begin{equation}
% P_i^{ir}=\psi_i\!\big(F_i^{ir}+\operatorname{Up}(P_{i+1}^{ir})\big),
% \qquad i\in\{1,\dots,L-1\}.
% \end{equation}
% This design preserves fine thermal edges at shallow layers while forming more stable semantic responses at deeper layers.

%%
%% The next two lines define the bibliography style to be used, and
%% the bibliography file.
\bibliographystyle{ACM-Reference-Format}
\bibliography{main}

\clearpage

\noindent\textbf{Additional details and supplementary results are provided in the appendices, organized as follows.}
\begin{itemize}
    \item \textbf{Appendix \ref{sec:appdx-A}}: Detailed implementation settings, including training configuration, model architecture, loss functions, and dataset setup.
    \item \textbf{Appendix \ref{sec:appdx-B}}: Additional qualitative comparisons on the LLVIP and M3FD datasets.
    \item \textbf{Appendix \ref{sec:appdx-C}}: Supplementary analytical experiments, including the roles of text, the efficiency--accuracy comparison of IR backbones, and degradation-aware analyses.
\end{itemize}

\appendix

\section{Implementation Details}
\label{sec:appdx-A}

This appendix provides the detailed implementation settings of the proposed framework, including the training configuration, model architecture, loss design, and dataset-specific setup. These details supplement the main paper and are intended to support reproducibility and a clearer understanding of the practical training protocol.

\subsection{Training Configuration}

Our model is built upon the YOLO framework.

\paragraph{Optimizer} We use the AdamW optimizer with weight decay set to 0.025 and momentum parameters $\beta_1=0.9$, $\beta_2=0.999$. The initial learning rate is set to $2\times10^{-3}$ and follows a linear decay schedule, gradually reducing the learning rate to 1\% of the peak ($lr\_factor=0.01$). Gradient clipping is applied with a maximum L2 norm of 10, and an exponential moving average (EMA) with momentum $1\times10^{-4}$ is used to stabilize training.

\paragraph{Differentiated Learning Rates} The CLIP model remains frozen during training, while all other learnable modules use the full learning rate ($lr\_mult=1.0$). Bias parameters, normalization layer parameters, and learnable logit scales are exempt from weight decay.

\paragraph{Training Schedule} All models are trained on a single H20 GPU with a batch size of 32. Validation is performed once per epoch, and the best checkpoint is selected based on mAP50. The final results are averaged over five independent runs.

\subsection{Model Architecture}

Table~\ref{tab:model-arch} summarizes the architectural configuration of each module. Configurations are consistent across all four datasets.

\begin{table*}[t!]
\centering
\caption{Configuration of model modules.}
\label{tab:model-arch}
\resizebox{\textwidth}{!}{
\begin{tabular}{ll}
\hline
Module & Configuration \\
\hline
RGB Backbone & CSPDarknet (P5, $d_f=0.33$, $w_f=0.5$), output channels [128, 256, 512] \\
IR Backbone & Proposed IR Backbone (LiteDCTGhostIRBackboneV2), output channels [64, 128, 256],\\
& $C_{base}=32$, $r_{freq}=0.5$, $r_{ghost}=2$, $r_{low\_freq\_init}=0.25$, with FPN \\
Text Backbone & CLIP ViT-B/32 (all parameters frozen) \\
Semantic-bridge-GuidedDynamicFusion &  Dynamic recalibration initialization $\alpha=0.5$, $\beta=0.5$ \\
Detection Head & YOLOWorldHead, $d_{embed}=512$, BN head \\
\hline
\end{tabular}
}
\end{table*}

\subsection{Loss Functions}

We adopt standard YOLO detection losses:

\begin{itemize}
\item \textbf{Classification loss:} Sigmoid binary cross-entropy, weight = 0.5
\item \textbf{Bounding box loss:} CIoU loss, weight = 7.5
\item \textbf{Distribution focal loss (DFL):} weight = 0.375
\end{itemize}

Label assignment is performed using BatchTaskAlignedAssigner with $top\_k=10$.

\subsection{Dataset Details}

We evaluate our method on four RGB-IR benchmark datasets:

\begin{itemize}
\item \textbf{FLIR-Aligned}: Contains 4 classes (car, person, bicycle, dog) with spatially aligned RGB-IR pairs and fixed train/val splits.
\item \textbf{LLVIP}: Night-time pedestrian surveillance dataset with 1 class (person), strictly aligned RGB-IR pairs.
\item \textbf{DroneVehicle}: Drone-view vehicle detection dataset with 5 classes (bus, car, freight\_car, truck, van). Original resolution $840\times712$, resized to $640\times640$ via letter-box padding during training.
\item \textbf{M3FD}: Multi-scenario RGB-IR dataset with 6 classes (People, Car, Bus, Motorcycle, Lamp, Truck) covering diverse lighting and weather conditions.
\end{itemize}

For each dataset, class-specific text prompts are provided to the frozen CLIP text encoder according to the YOLO-World protocol.

\section{Supplementary Qualitative Results}
\label{sec:appdx-B}

This appendix presents additional qualitative comparisons on the LLVIP and M3FD datasets. By examining representative challenging scenes, we further illustrate the advantages of the proposed method in multispectral perception, particularly in terms of detection completeness, localization accuracy, and robustness under adverse illumination, occlusion, and cluttered backgrounds.

\subsection{Visual comparison on LLVIP dataset}

\begin{figure*}[t!]
  \centering
  \includegraphics[width=0.7\textwidth,height=8.5cm]{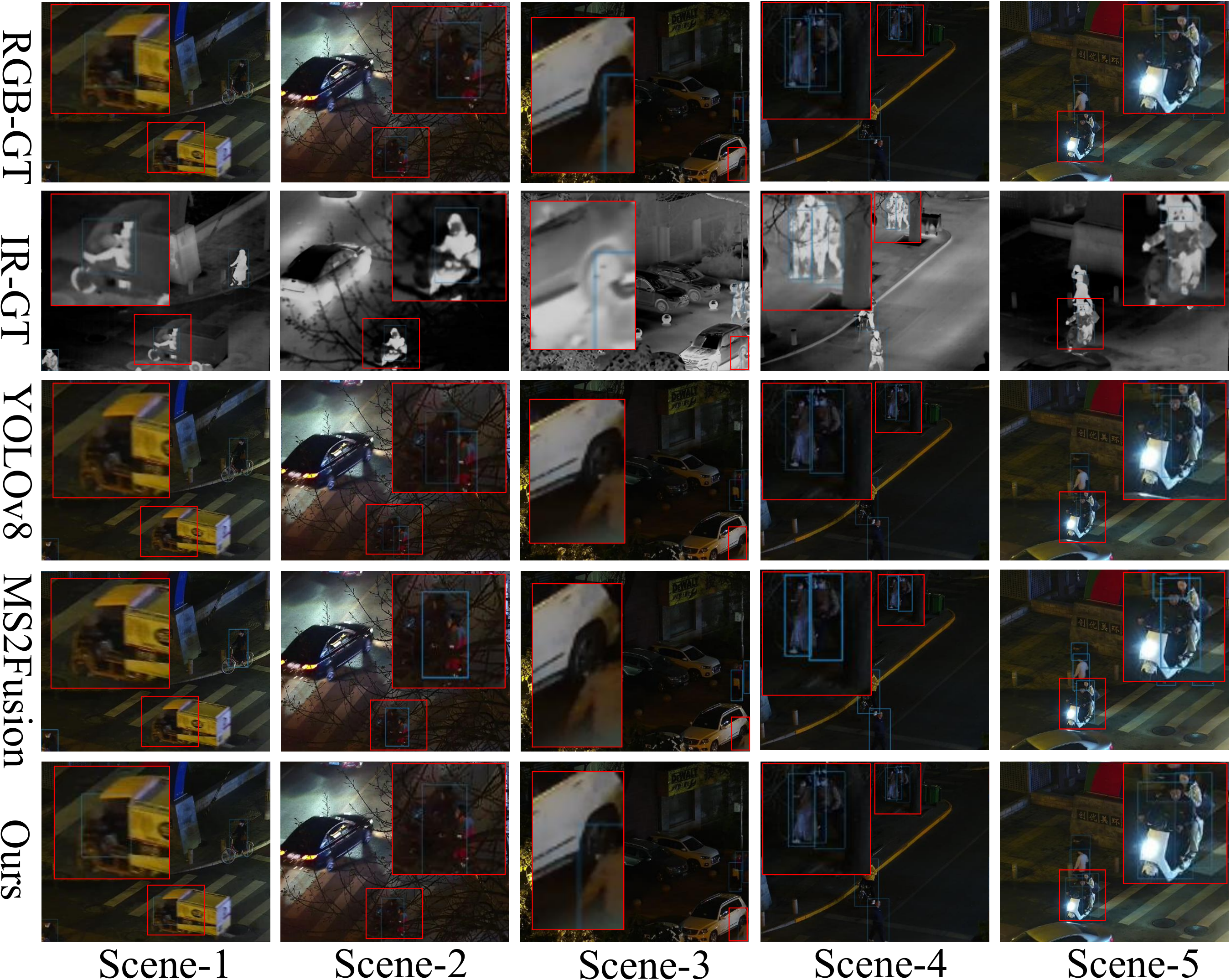}
  \caption{Qualitative comparison on five representative scenes from the LLVIP dataset. }
  \Description{A woman and a girl in white dresses sit in an open car.}
  \label{fig:visual-llvip}
\end{figure*}

Figure~\ref{fig:visual-llvip} presents qualitative detection results across five challenging nighttime scenes using different methods: YOLOv8, MS2Fusion, and our proposed approach. The first two rows show ground-truth annotations for RGB images (RGB-GT) and infrared images (IR-GT), respectively. The subsequent rows display detection outputs from YOLOv8, MS2Fusion, and our method.

Observations from the figure include:

\begin{itemize}
    \item \textbf{Complementarity of RGB and IR modalities:} RGB images alone suffer from low contrast and occlusion in nighttime scenes (e.g., Scenes 2 and 4), leading to missed detections by single-modality methods. Infrared annotations (IR-GT) highlight these regions clearly, demonstrating the value of leveraging both modalities.

    \item \textbf{Detection completeness:} Compared with YOLOv8 and MS2Fusion, our method consistently detects all annotated objects, including small and partially occluded targets (e.g., Scene-2 and Scene-5). This indicates that semantic-bridge guided dynamic fusion effectively integrates information from RGB and IR, improving detection coverage.

    \item \textbf{Localization accuracy:} Zoomed-in regions illustrate that our method provides more precise bounding boxes than previous methods. MS2Fusion occasionally misaligns or misses small objects under complex lighting (Scene-3), whereas our approach maintains accurate localization across all scenes.

    \item \textbf{Robustness to visual noise:} Infrared ground truth (IR-GT) emphasizes object saliency under poor illumination. While MS2Fusion sometimes produces false positives in bright regions, our method reduces spurious detections by exploiting consensus and discrepancy patterns between RGB and IR modalities.
\end{itemize}

\subsection{Visual comparison on M3FD dataset}

\begin{figure*}[t!]
  \centering
  \includegraphics[width=0.7\textwidth,height=8.5cm]{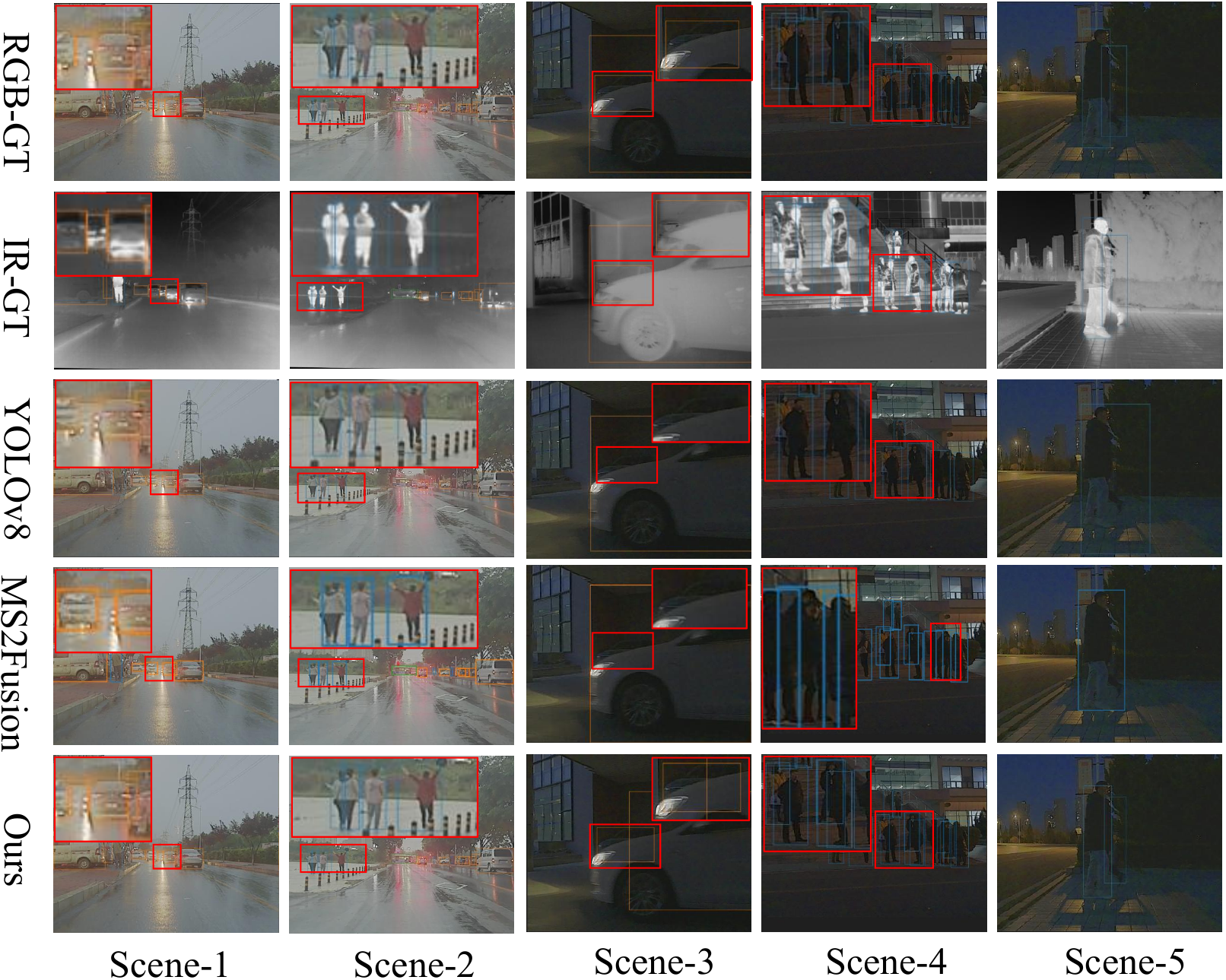}
  \caption{Qualitative comparison on five representative scenes from the M3FD dataset. }
  \Description{A woman and a girl in white dresses sit in an open car.}
  \label{fig:visual-m3fd}
\end{figure*}

Figure~\ref{fig:visual-m3fd} shows qualitative detection results on five challenging scenarios from the M3FD dataset. The first two rows correspond to ground-truth annotations for RGB images (RGB-GT) and infrared images (IR-GT), respectively. The subsequent rows display detection outputs from YOLOv8, MS2Fusion, and our proposed method.

Key observations include:

\begin{itemize}
    \item \textbf{Multimodal complementarity:} RGB images in adverse lighting or weather conditions (e.g., Scenes 1 and 2) suffer from low contrast and occlusion. Infrared annotations (IR-GT) clearly highlight pedestrians, vehicles, and other small objects that are challenging to detect in RGB alone, demonstrating the importance of multispectral fusion.

    \item \textbf{Detection coverage:} Our method consistently detects all annotated targets across diverse scenarios, including small, partially occluded, or distant objects. YOLOv8 and MS2Fusion occasionally miss peripheral or low-contrast objects (e.g., Scene-4), whereas our approach maintains high completeness.

    \item \textbf{Bounding-box precision:} Zoom-in regions indicate that our method produces tighter and more accurate bounding boxes compared with previous methods, reducing misalignment especially in cluttered scenes (Scene-3 and Scene-4).

    \item \textbf{Robustness to environmental noise:} The fusion strategy in our method leverages consensus and discrepancy patterns between RGB and IR features, which helps suppress false positives in reflective or highly illuminated areas, as observed in Scene-5.
\end{itemize}

Overall, the qualitative results confirm that semantic-aware RGB-IR fusion improves detection reliability, localization accuracy, and completeness in complex multi-scenario environments, outperforming both single-modality and prior multimodal methods.

\section{Supplementary Analytical Experiments}
\label{sec:appdx-C}

This appendix provides further analytical experiments to complement the main paper. Specifically, we further analyze the two roles of text in the proposed framework (Section~\ref{sec:textrole-appdx}), supplement the efficiency--accuracy comparison of infrared backbones (Section~\ref{sec:IR-backbones-appdx}), and present more detailed degradation-aware analyses from both instance-level and population-level perspectives (Section~\ref{sec:Degradation-Experiment-appdx}). Together, these results provide further evidence for the effectiveness, efficiency, and robustness of the proposed design.

\subsection{Analysis of the Roles of Text}
\label{sec:textrole-appdx}

We further conduct a detailed analysis of the roles of text on the FLIR dataset, as summarized in Table~\ref{tab:text_role_analysis}.

\paragraph{Text as a Semantic Bridge.}
Under the \emph{Semantic Bridge} paradigm, we consider two classic multispectral fusion schemes, namely \emph{Vanilla Direct RGB--IR Fusion} and \emph{Conditional Prompt Fusion}. The former is representative of direct cross-modal fusion pipelines\footnote{Representative methods include UniRGB-IR: \url{https://github.com/PoTsui99/UniRGB-IR} and DroneShip: \url{https://github.com/luting-hnu/DroneShip}.}, while the latter injects text as an additional conditional prompt for multimodal interaction\footnote{Representative methods include OmniFuse: \url{https://github.com/HaoZhang1018/OmniFuse}, SLGNet: \url{https://github.com/Xiantai01/SLGNet}, and ControlFusion: \url{https://github.com/Linfeng-Tang/ControlFusion}.}. For fair comparison, we re-implement both paradigms within our proposed framework. The implementation details are provided in Section~\ref{sec:Implementation-Fusion-Paradigms}.

\begin{figure}[h]
    \centering
    \begin{subfigure}[t]{0.48\linewidth}
        \centering
        \includegraphics[width=\linewidth]{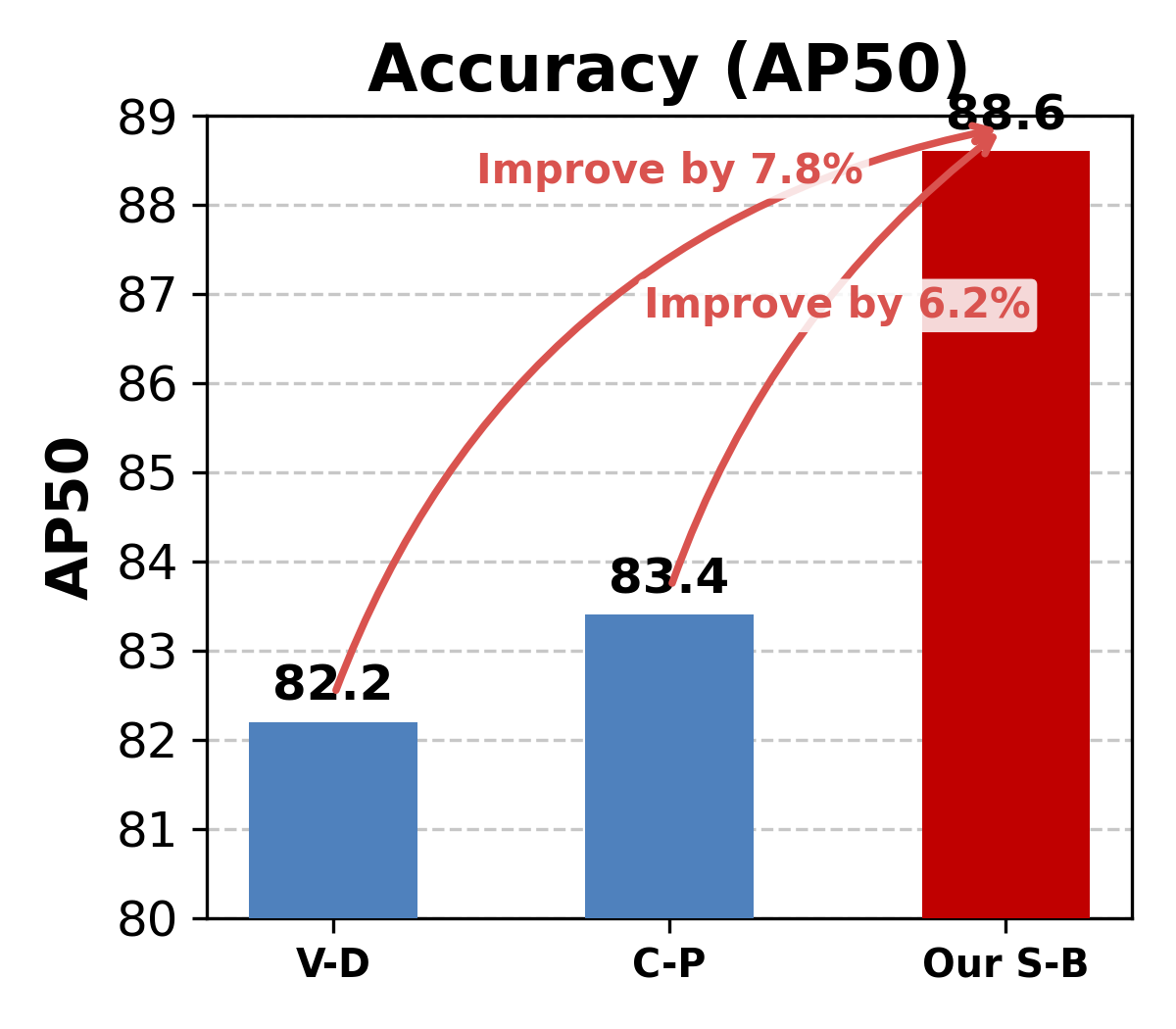}
        \caption{Comparison of different fusion paradigms in terms of AP$_{50}$.}
        \label{fig:sub_a-appdx}
    \end{subfigure}
    \hfill
    \begin{subfigure}[t]{0.48\linewidth}
        \centering
        \includegraphics[width=\linewidth]{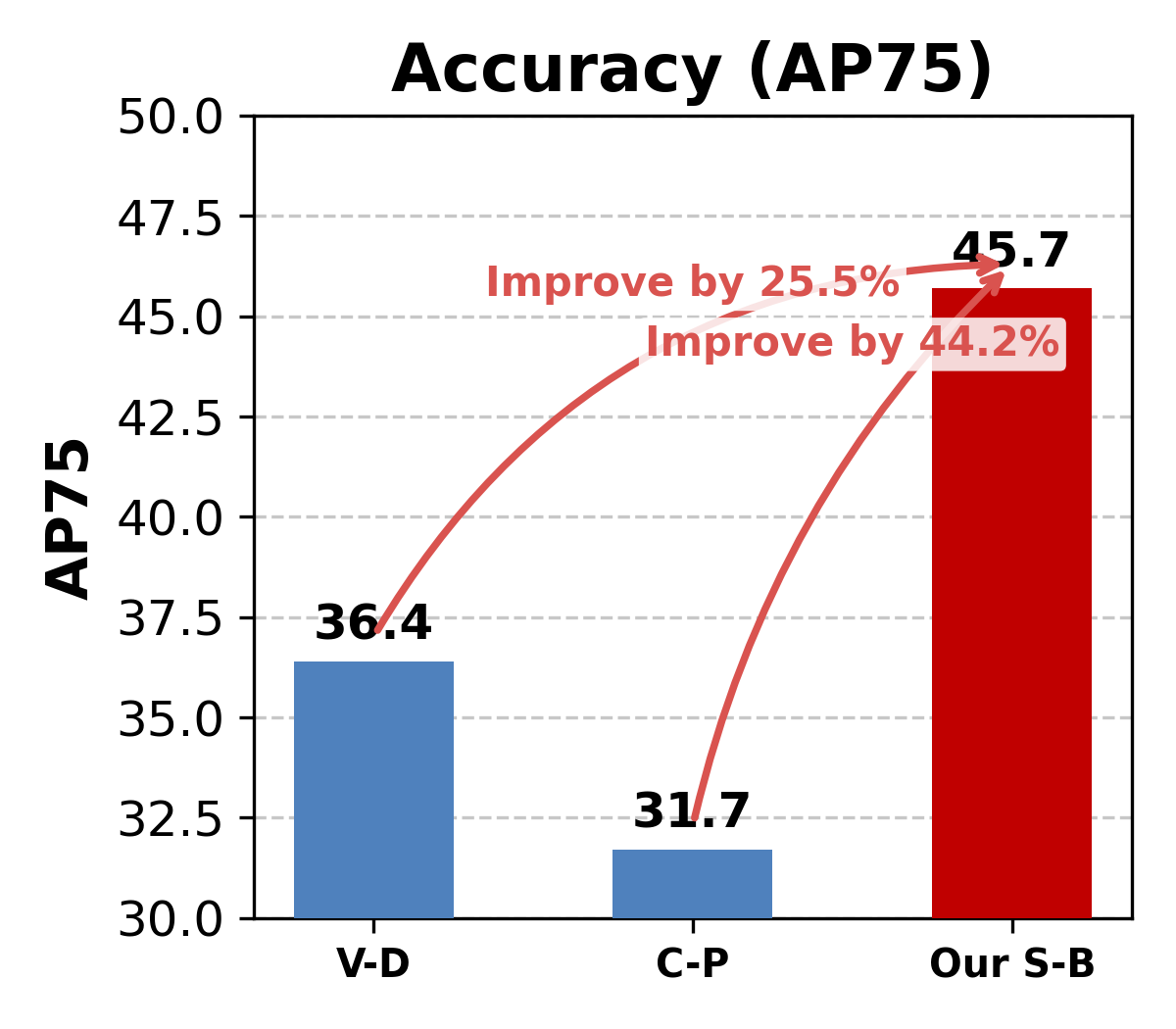}
        \caption{Comparison of different fusion paradigms in terms of AP$_{75}$.}
        \label{fig:sub_b-appdx}
    \end{subfigure}\\

    \begin{subfigure}[t]{0.48\linewidth}
        \centering
        \includegraphics[width=\linewidth]{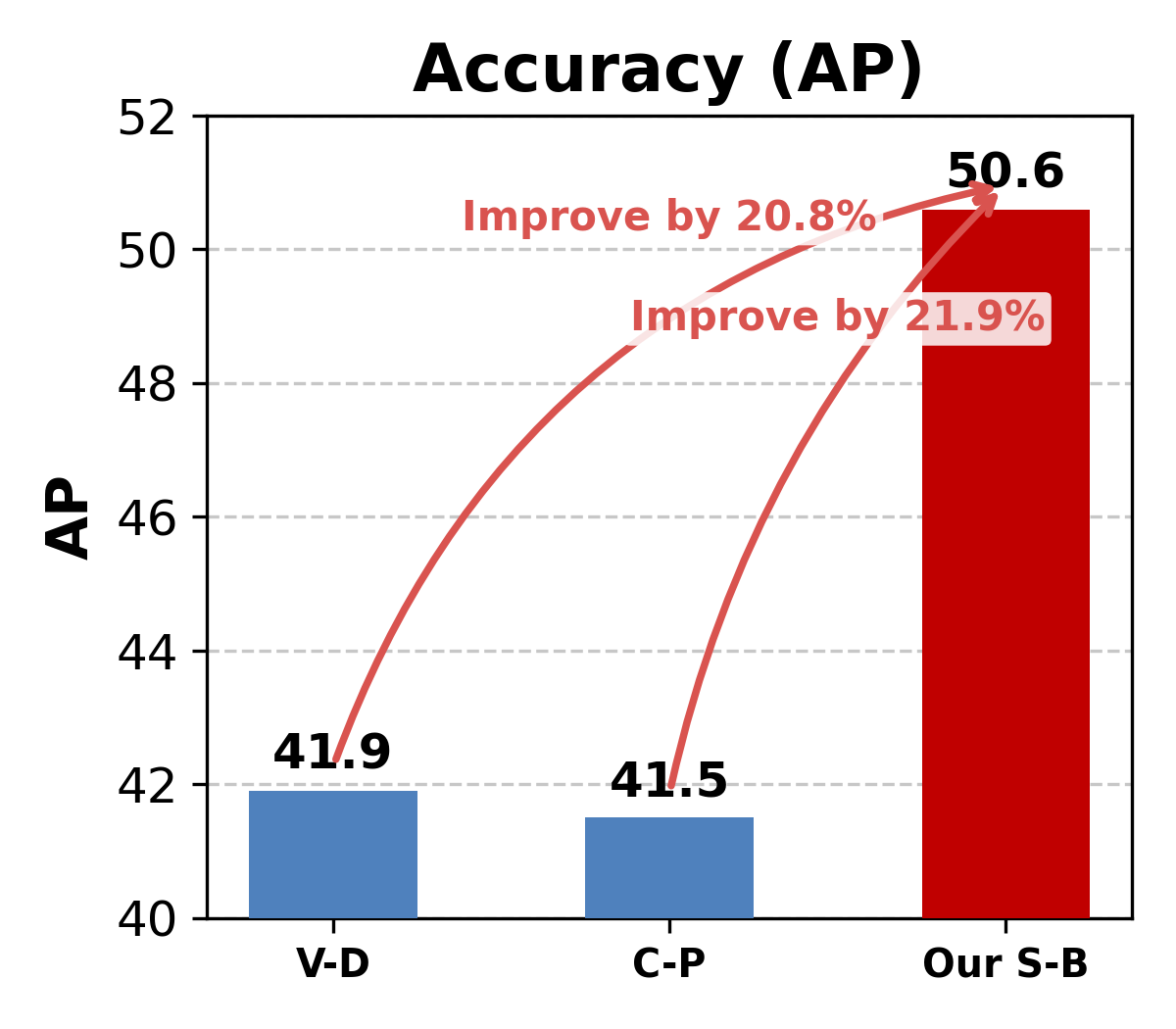}
        \caption{Comparison of different fusion paradigms in terms of AP.}
        \label{fig:sub_c-appdx}
    \end{subfigure}
    \hfill
    \begin{subfigure}[t]{0.48\linewidth}
        \centering
        \includegraphics[width=\linewidth]{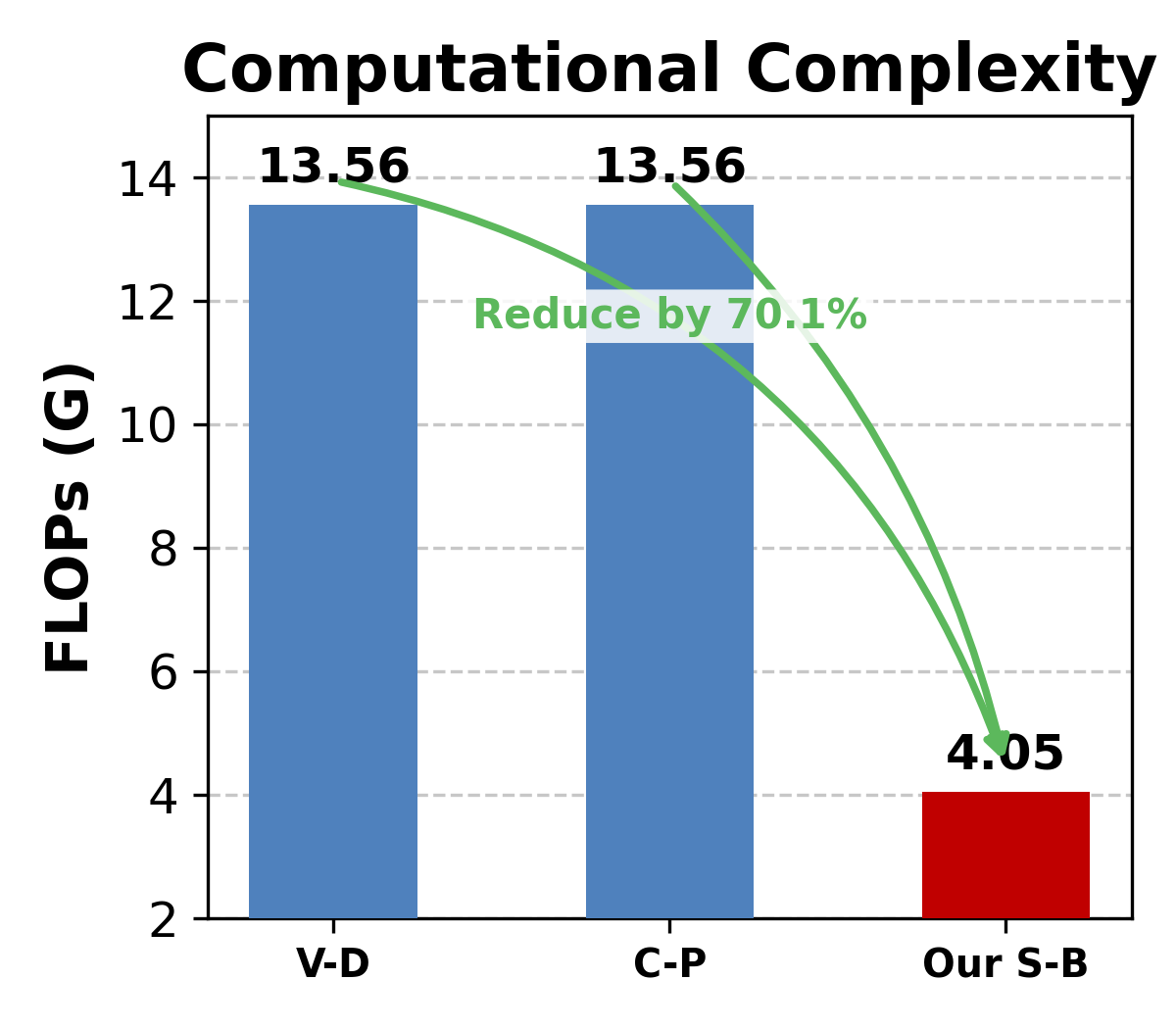}
        \caption{Comparison of different fusion paradigms in terms of computational complexity.}
        \label{fig:sub_d-appdx}
    \end{subfigure}    
    \caption{Comparison of different fusion paradigms on the FLIR dataset in terms of computational complexity and detection performance. V-D, C-P, and Our S-B denote Vanilla Direct Fusion, Conditional Prompt Fusion, and our Semantic Bridge Fusion, respectively.}  
    \label{fig:appd-comp-appdx}
\end{figure}

As further visualized in Fig.~\ref{fig:appd-comp-appdx}, our method consistently outperforms both reproduced paradigms in terms of both accuracy and computational efficiency. Specifically, compared with the two comparison paradigms, our method improves AP from 41.7 to 50.6, corresponding to a relative gain of 21.3\%, while reducing FLOPs from 13.56G to 4.05G, yielding a reduction of 70.1\%. These results indicate that \emph{text as a Semantic Bridge} not only strengthens semantically aligned cross-modal interaction, but also substantially reduces the computational burden of multispectral fusion.

The large FLOPs reduction may appear counter-intuitive at first glance. However, within our framework, lightweight text features act as a bottleneck, significantly reducing the intermediate computational cost. A detailed derivation is provided in section \ref{sec:text_Semantic_Bridge}.

\paragraph{Text as a Semantic Anchor.}
For the \emph{Semantic Anchor} role, we consider two baselines: removing text embeddings entirely, and \emph{Fixed Random Text Embedding}, where each category is assigned a fixed random vector as its embedding. The results show that text serving as a semantic anchor provides stable category-level guidance, thereby improving the robustness and semantic consistency of detection.

\subsection{Implementation Details of the Compared Fusion Paradigms.}
\label{sec:Implementation-Fusion-Paradigms}
For fair comparison, we re-implement two representative fusion paradigms within our unified framework, namely \emph{Vanilla Direct RGB--IR Fusion} and \emph{Conditional Prompt Fusion}.

\paragraph{Vanilla Direct RGB--IR Fusion.}
This variant removes all text-guided semantic reasoning and replaces the proposed semantic bridge with direct spatial cross-attention between RGB and IR features, aiming to examine whether pure visual cross-modal interaction is sufficient without text-anchored semantic decomposition.

Specifically, the IR and RGB features are first projected into a shared \(d_k\)-dimensional space using \(1\times1\) convolutions:
\begin{equation}
Q = W_Q * X_{ir} \in \mathbb{R}^{B \times d_k \times HW}, \qquad
K = W_K * X_{rgb} \in \mathbb{R}^{B \times d_k \times HW}.
\end{equation}
A full spatial attention matrix is then computed as
\begin{equation}
A = \sigma\left(\frac{Q^\top K}{\sqrt{d_k}}\right) \in \mathbb{R}^{B \times HW \times HW},
\end{equation}
where \(\sigma(\cdot)\) denotes the sigmoid function.

\paragraph{Conditional Prompt Fusion.}
This variant introduces text through a FiLM-style conditioning mechanism. Following the prompt modulation design in ControlFusion, text globally modulates the fusion response via channel-wise affine transformation.

Concretely, the same IR--RGB spatial cross-attention as above is first used to obtain the spatial response map \(R\). Meanwhile, the text embedding
\begin{equation}
T \in \mathbb{R}^{B \times N_t \times d_t}
\end{equation}
is fed into a two-layer MLP with structure \(d_t \rightarrow d_t/2 \rightarrow 2C_{ir}\) and GELU activation to generate channel-wise affine parameters:
\begin{equation}
[\gamma_p,\beta_p] = \mathrm{MLP}(T), \qquad
\gamma_p,\beta_p \in \mathbb{R}^{B \times C_{ir} \times 1 \times 1}.
\end{equation}
The modulation is applied as
\begin{equation}
F_{\mathrm{mod}} = (1+\gamma_p)\odot R + \beta_p.
\end{equation}

\subsection{Analysis of the Computational Efficiency of Text as a Semantic Bridge}
\label{sec:text_Semantic_Bridge}
To explain why \emph{text as a Semantic Bridge} can simultaneously improve detection performance and reduce FLOPs, we provide a simplified analysis of the dominant cost of cross-modal interaction. Here we only consider the core matrix multiplications in the fusion module, while ignoring secondary operations such as linear projections, normalization, and activation, since they do not affect the leading-order complexity.

Let the spatial size of the feature map be \(H \times W\), and define
\begin{equation}
N = H \times W,
\end{equation}
where \(N\) denotes the number of visual tokens. Let the RGB and IR features be
\begin{equation}
X_{rgb}, X_{ir} \in \mathbb{R}^{N \times C},
\end{equation}
and let the text feature be
\begin{equation}
T \in \mathbb{R}^{M \times C}.
\end{equation}
Here, \(M\) denotes the number of categories rather than the number of generic text tokens. In the datasets considered in this paper, the category numbers are \(4\), \(1\), \(5\), and \(6\) for FLIR, LLVIP, DroneVehicle, and M3FD, respectively. More generally, in object detection, the number of categories typically satisfies
\begin{equation}
M \ll N.
\end{equation}

For a matrix multiplication with
\begin{equation}
A \in \mathbb{R}^{a \times b}, \qquad B \in \mathbb{R}^{b \times c},
\end{equation}
its computational cost is counted as
\begin{equation}
\mathrm{FLOPs}(AB) = 2abc.
\end{equation}

For direct RGB--IR interaction, if a global affinity matrix is constructed directly between the two visual modalities, we have
\begin{equation}
A_{dir} = X_{ir} X_{rgb}^{\top} \in \mathbb{R}^{N \times N},
\end{equation}
which is further used for feature aggregation. The dominant computational cost is therefore
\begin{equation}
\mathrm{FLOPs}_{dir} \approx 4N^{2}C.
\end{equation}

In contrast, in our method, text serves as a category-level semantic bridge and factorizes the direct RGB--IR interaction into two steps:
\begin{equation}
A_{ir \rightarrow t} = X_{ir} T^{\top} \in \mathbb{R}^{N \times M},
\qquad
A_{t \rightarrow rgb} = T X_{rgb}^{\top} \in \mathbb{R}^{M \times N}.
\end{equation}
Accordingly, the dominant computational cost becomes
\begin{equation}
\mathrm{FLOPs}_{bridge} \approx 2NCM + 2MCN = 4NMC.
\end{equation}

The resulting complexity ratio is
\begin{equation}
\frac{\mathrm{FLOPs}_{bridge}}{\mathrm{FLOPs}_{dir}}
=
\frac{4NMC}{4N^{2}C}
=
\frac{M}{N}.
\end{equation}

This result shows that direct RGB--IR interaction has complexity \(O(N^{2}C)\), whereas the Semantic Bridge formulation reduces it to \(O(NMC)\). Since \(M\) denotes the category number and is typically much smaller than the number of visual tokens \(N\), our design transforms the originally expensive dense visual--visual interaction into a low-rank interaction between visual features and a small set of category-level semantic anchors, thereby substantially reducing the computational cost of the fusion module.

\subsection{Efficiency and accuracy comparison of IR backbones}
\label{sec:IR-backbones-appdx}

We analyze the efficiency and accuracy of the IR backbone on the FLIR dataset (image resolution $640\times640$) (Table~\ref{tab:ir_backbone}). 
\textbf{Compared with the classical feature extraction backbone ResNet50}, our IR backbone contains only $1.05$M parameters, a reduction of approximately $95\%$ compared to ResNet50's $24.20$M, and requires $6.49$G FLOPs, which is about $81\%$ lower than ResNet50. 
In terms of inference efficiency, our IR backbone achieves a latency of $3.30$ ms per image, corresponding to $302.7$ FPS, representing improvements of approximately $30\%$ and $42\%$ over ResNet50, respectively. 

\begin{table}[h]
\centering
\caption{Efficiency and accuracy comparison of IR backbones on the FLIR dataset (image size $640\times640$).}
\label{tab:ir_backbone}
\resizebox{\columnwidth}{!}{
\begin{tabular}{lcccccc}
\toprule
IR Backbone & Params (M) & FLOPs (G) & Latency (ms) & FPS & mAP50 & mAP \\
\midrule
ResNet50 & 24.20 & 34.18 & 4.69 & 213.3 & 0.853 & 0.453 \\
Our IR backbone & 1.05 & 6.49 & 3.30 & 302.7 & 0.886 & 0.504 \\
\bottomrule
\end{tabular}
}
\end{table}

Notably, the lightweight design does not compromise detection accuracy: our IR backbone attains $0.886$ mAP50 and $0.504$ mAP, outperforming ResNet50 ($0.853$ mAP50 and $0.453$ mAP). 
These results demonstrate that the proposed IR backbone significantly improves computational efficiency and real-time performance while maintaining high precision, providing \textbf{effective support for real-time multispectral object detection}.

\subsection{Degradation Experiment Analysis}
\label{sec:Degradation-Experiment-appdx}

We further conduct a detailed analysis of two degradation experiments to more fully validate the effectiveness and complementarity of the proposed \textbf{Bi-Support pattern}, namely instance-level progressive degradation (Section~\ref{sec:instance-appdx}) and population-level progressive degradation (Section~\ref{sec:population-appdx}). Both experiments are conducted on the LLVIP dataset, where synthetic progressive degradation is applied only to the RGB modality while the infrared modality remains unchanged. We further analyze the variation trends of the activation values of $M_{\text{cons}}$ and $M_{\text{dis}}$ under increasing degradation levels.

\paragraph{Degradation Setup.} We apply a composite degradation strategy that more closely reflects real-world conditions to perturb the RGB images:

\begin{itemize}
\item \textbf{Brightness scaling:} Multiply pixel intensities by a factor $\gamma$ to simulate under- or over-exposure.
\item \textbf{Gaussian blur:} Apply a $k\times k$ Gaussian kernel to simulate defocus or atmospheric scattering.
\item \textbf{Additive Gaussian noise:} Add zero-mean Gaussian noise with standard deviation $\sigma$ to simulate sensor noise.
\end{itemize}

A total of \textbf{discrete levels} are defined, and their parameter settings are summarized in Table~\ref{tab:degradation}. In the subsequent population-level analysis, four representative levels are selected from the discrete levels and referred to as Degradation-1, Degradation-2, Degradation-3, and Degradation-4 (see Fig. \ref{fig:degrade-apped}).

\begin{table}[h]
\centering
\caption{Parameters for discrete degradation levels. Four representative levels are selected for population-level statistics.}
\label{tab:degradation}
\begin{tabular}{ccccc}
\hline
Level & Brightness ($\gamma$) & Sharpen & Blur ($k$) & Noise ($\sigma$) \\
\hline
0 & No modification & -- & -- & -- \\
+1 & 0.70 & -- & 5 & 8 \\
+2 & 0.55 & -- & 9 & 15 \\
+3 & 0.40 & -- & 15 & 22 \\
+4 & 0.28 & -- & 21 & 32 \\
+5 & 0.18 & -- & 29 & 42 \\
+6 & 0.12 & -- & 37 & 55 \\
+7 & 0.07 & -- & 45 & 65 \\
+8 & 0.04 & -- & 55 & 80 \\
+9 & 0.02 & -- & 65 & 95 \\
+10 & 0.01 & -- & 81 & 110 \\
\hline
\end{tabular}
\end{table}

\subsection{Instance-Level Analysis}
\label{sec:instance-appdx}

To provide a finer-grained understanding of the proposed Bi-Support pattern under different RGB quality conditions, we conduct an instance-level analysis on individual image pairs. 

\paragraph{Foreground Activation Evaluation Protocol.}

For each LLVIP image pair, the infrared modality is kept unchanged, while the RGB modality is progressively degraded. This experiment aims to examine how the pixel-wise foreground activation patterns of $M_{\text{cons}}$ and $M_{\text{dis}}$ evolve as the RGB quality deteriorates. Here, the \textit{Activation} specifically refers to the average activation response within the ground-truth box region, which is formally defined below as the instance-level response. The computation is given as follows.

The two semantic response maps are defined as Eq. \ref{eq:bi-support}, where $A_{\text{ir}}$ and $A_{\text{rgb}}$ denote the normalized activation maps of the infrared and RGB modalities, respectively.

First, each response map is normalized independently using min-max normalization:
\begin{equation}
\tilde{M}(p)=\frac{M(p)-\min_p M(p)}{\max_p M(p)-\min_p M(p)+\epsilon},
\end{equation}
where $\epsilon=10^{-6}$ is used to avoid division by zero, and $p$ denotes the spatial position index.

Based on the normalized response maps, we construct IR-weighted feature maps ($X_{\text{ir}}$) as
\begin{equation}
F_{\text{cons}} = X_{\text{ir}} \odot \tilde{M}_{\text{cons}}, \qquad
F_{\text{dis}} = X_{\text{ir}} \odot \tilde{M}_{\text{dis}}.
\end{equation}

For each degradation level $l$, we obtain $F_{\text{cons}}^{(l)}$ and $F_{\text{dis}}^{(l)}$. To enable fair comparison across degradation levels, each type of feature map is further globally normalized over all levels, yielding $\hat{F}_{\text{cons}}^{(l)}$ and $\hat{F}_{\text{dis}}^{(l)}$.

Finally, given a ground-truth bounding box $\mathcal{B}=[x_1,y_1,x_2,y_2]$, we map it onto the feature space with stride $s=8$ to obtain the corresponding region $\mathcal{B}_f$. The instance-level responses are then defined as the average activations within the target region:
\begin{equation}
\bar{R}_{\text{cons}}^{(l)}=
\frac{1}{|\mathcal{B}_f|}\sum_{p\in \mathcal{B}_f}\hat{F}_{\text{cons}}^{(l)}(p), \qquad
\bar{R}_{\text{dis}}^{(l)}=
\frac{1}{|\mathcal{B}_f|}\sum_{p\in \mathcal{B}_f}\hat{F}_{\text{dis}}^{(l)}(p).
\end{equation}

These metrics characterize how the consensus response and discrepancy response evolve within the target region as the RGB modality is progressively degraded.

\begin{figure}[t!]
    \centering
    \begin{subfigure}[t]{\linewidth}
        \centering
        \includegraphics[width=0.57\textwidth,height=5cm]{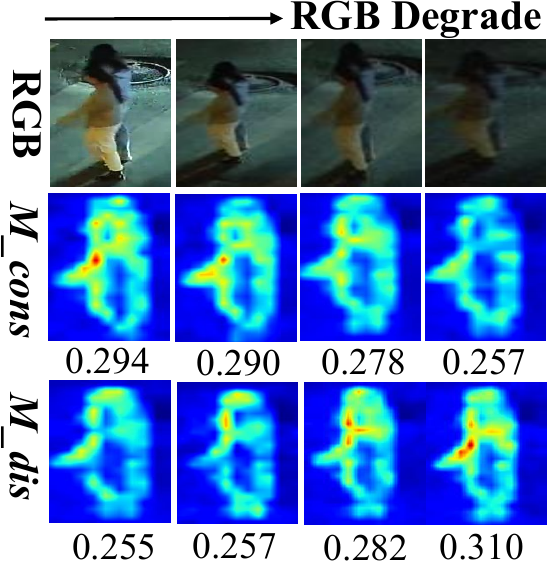}
        \caption{Additional instance-level example 1. As the RGB modality is progressively degraded, the consensus response within the target region gradually decreases, while the discrepancy response correspondingly increases.}
        \label{fig:sub_a-degrade-appdx}
    \end{subfigure}
    \begin{subfigure}[t]{\linewidth}
        \centering
        \includegraphics[width=0.57\textwidth,height=5cm]{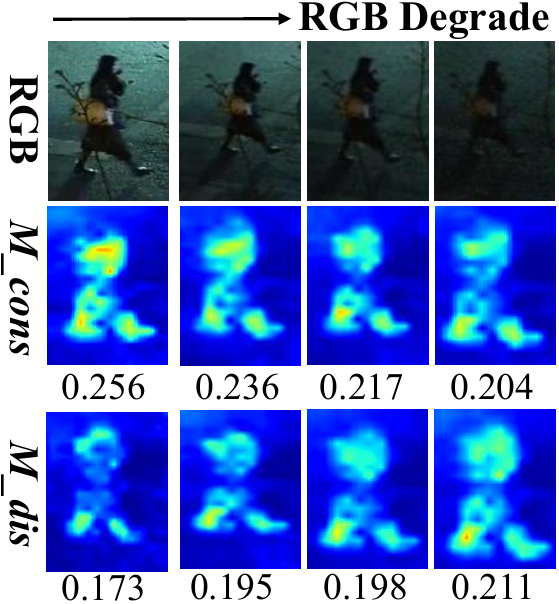}
        \caption{Additional instance-level example 2. A similar trend is observed on another sample, where RGB degradation suppresses consensus support but strengthens discrepancy support in the target area.}
        \label{fig:sub_b-degrade-appdx}
    \end{subfigure}
    \caption{Visualization of responses of {consensus support} $M_{cons}$ and {discrepancy support} $M_{dis}$ under progressively degraded RGB inputs on typical samples.}  
    \label{fig:appd-comp-degrade-appdx}
\end{figure}

\paragraph{Additional Instance-Level Examples.}

Beyond Fig.~\ref{fig:degrade}, we provide more instance-level examples to further verify the generality of the observed trend (Fig. \ref{fig:appd-comp-degrade-appdx}). As the RGB modality becomes increasingly degraded, the response of consensus support $M_{\text{cons}}$ consistently weakens inside the target region, whereas the response of discrepancy support $M_{\text{dis}}$ becomes progressively stronger. This behavior is quantitatively reflected by the average responses within the GT region. In the first example, the mean consensus response decreases from 0.294 to 0.257, while the mean discrepancy response increases from 0.255 to 0.310. A similar pattern is observed in the second example, where the consensus response drops from 0.256 to 0.204 and the discrepancy response rises from 0.173 to 0.211. These results indicate that, as RGB information becomes less reliable, the model gradually shifts from cross-modal consensus cues to discrepancy-aware IR-dominant cues, demonstrating the adaptive behavior of the proposed Bi-Support pattern at the instance level.

\subsection{Population-Level Analysis}
\label{sec:population-appdx}

To verify that observed patterns are not due to single images, we extend the analysis to a set of LLVIP images.

\paragraph{Foreground Activation Evaluation Protocol.}
To verify the statistical stability of the instance-level observations, we analyze the population-level migration patterns of consensus support and discrepancy support over the entire LLVIP training set. For each image $i$ at each degradation level $l$, we compute the feature maps $F_{\text{cons},i}^{(l)}$ and $F_{\text{dis},i}^{(l)}$. The target-region mask $\mathcal{G}_i$ is defined as the union of all annotated bounding boxes projected onto the feature space with stride $s=8$:
\begin{equation}
\mathcal{G}_i=\bigcup_{j=1}^{N_i}\mathcal{B}_{i,j}^{(f)},
\end{equation}
where $N_i$ denotes the number of annotated instances in image $i$.

To ensure comparability across degradation levels, we perform global min-max normalization over all levels for each image and each feature type. Taking the consensus feature map as an example:
\begin{equation}
v_{\min,i}^{\text{cons}}=\min_{l}\min_{p}F_{\text{cons},i}^{(l)}(p), \qquad
v_{\max,i}^{\text{cons}}=\max_{l}\max_{p}F_{\text{cons},i}^{(l)}(p),
\end{equation}
\begin{equation}
\hat{F}_{\text{cons},i}^{(l)}(p)=
\frac{F_{\text{cons},i}^{(l)}(p)-v_{\min,i}^{\text{cons}}}
{v_{\max,i}^{\text{cons}}-v_{\min,i}^{\text{cons}}},
\end{equation}
where $p$ denotes the spatial position index. The same procedure is applied to $F_{\text{dis},i}^{(l)}$.

We then define the Normalized Mean Response Proportion (NMRP) to measure the average response over the target and background regions. For the consensus feature map, the GT-region and background-region NMRPs are given by
\begin{equation}
\mathrm{NMRP}_{\mathrm{GT}}^{\mathrm{cons}}(i,l)=
\frac{1}{|\mathcal{G}_i|}
\sum_{p\in\mathcal{G}_i}\hat{F}_{\text{cons},i}^{(l)}(p),
\end{equation}
\begin{equation}
\mathrm{NMRP}_{\mathrm{BG}}^{\mathrm{cons}}(i,l)=
\frac{1}{|\bar{\mathcal{G}}_i|}
\sum_{p\in\bar{\mathcal{G}}_i}\hat{F}_{\text{cons},i}^{(l)}(p),
\end{equation}
where $\bar{\mathcal{G}}_i$ denotes the background region. Analogously, we obtain $\mathrm{NMRP}_{\mathrm{GT}}^{\mathrm{dis}}(i,l)$ and $\mathrm{NMRP}_{\mathrm{BG}}^{\mathrm{dis}}(i,l)$ for the discrepancy feature map.

Finally, for each degradation level $l$, the population-level migration metrics are computed by averaging NMRP over all $N$ training images:
\begin{equation}
\begin{aligned}
\overline{\mathrm{NMRP}}_{\mathrm{GT}}^{\mathrm{cons}}(l)
&=\frac{1}{N}\sum_{i=1}^{N}\mathrm{NMRP}_{\mathrm{GT}}^{\mathrm{cons}}(i,l), \\
\overline{\mathrm{NMRP}}_{\mathrm{GT}}^{\mathrm{dis}}(l)
&=\frac{1}{N}\sum_{i=1}^{N}\mathrm{NMRP}_{\mathrm{GT}}^{\mathrm{dis}}(i,l).
\end{aligned}
\end{equation}
The background-region metrics are computed in the same manner. These measures characterize the population-level migration trends of consensus and discrepancy support in both target and background regions as the RGB modality is progressively degraded.

\begin{table}[h]
\centering
\caption{Foreground occupancy statistics within GT boxes on the LLVIP training set.}
\label{tab:gt_occupancy}
\begin{tabular}{lc}
\hline
Metric & Value \\
\hline
Number of GT boxes & 34,130 \\
Mean & 0.4112 \\
Median & 0.4051 \\
Standard deviation & 0.1447 \\
Minimum & 0.0181 \\
Maximum & 0.9601 \\
25th percentile & 0.3083 \\
75th percentile & 0.5069 \\
\hline
\end{tabular}
\end{table}

Additionally, we further estimate the true foreground occupancy ratio of each GT box to directly compare the foreground captured by the Bi-Support pattern against the actual foreground. Specifically, for each annotated box in the LLVIP training set, we crop the corresponding infrared region and perform foreground segmentation on the grayscale crop using Otsu thresholding\footnote{\url{https://zeal-up.github.io/2023/09/27/cv/otsu_method/}}. The foreground occupancy ratio is defined as
\begin{equation}
r=\frac{|\{p:\mathrm{binary}(p)=1\}|}{W_{\mathrm{box}}\times H_{\mathrm{box}}},
\end{equation}
where $W_{\mathrm{box}}$ and $H_{\mathrm{box}}$ denote the width and height of the GT box, respectively. The statistics are summarized in Table~\ref{tab:gt_occupancy}. On average, approximately $41.1\%$ of each GT box is occupied by the pedestrian, while the remaining $58.9\%$ corresponds to background. This indicates that the NMRP measured over GT regions inherently reflects a mixture of target response and local background response.

\begin{figure}[H]
  \centering
  \includegraphics[width=0.5\textwidth,height=5cm]{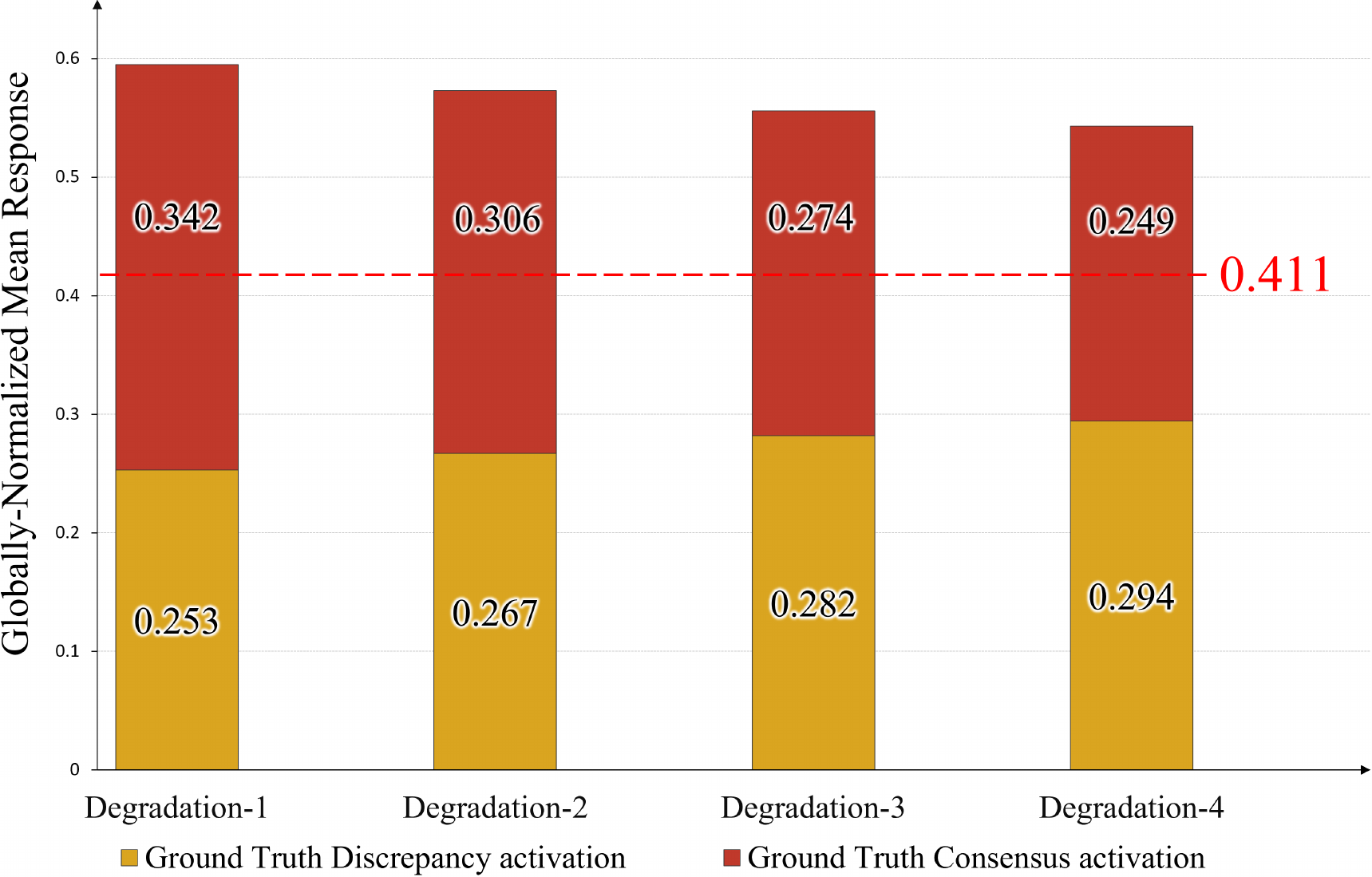}
  \vspace{-0.3cm}
  \caption{Population-level trends of consensus and discrepancy support across RGB degradation levels on LLVIP. The bars show the globally normalized mean responses in the GT regions, and the red dashed line denotes the average pedestrian occupancy ratio within GT boxes ($41.1\%$). With increasing RGB degradation, consensus support in the GT region decreases, whereas discrepancy support increases, indicating a progressive migration of discriminative evidence.}
  \label{fig:degrade-apped}
\end{figure}

\paragraph{Interpretation of Population-Level Migration.}
Building on the instance-level evidence in Fig.~\ref{fig:degrade-200}, the population-level statistics provide a complementary interpretation of why bi-support modeling is effective. A key observation (Fig.~\ref{fig:degrade-apped}) is that the GT-region consensus response alone already accounts for a substantial portion of the actual pedestrian occupancy within each GT box, whose dataset-level mean is approximately $41.1\%$. This indicates that consensus support captures a large amount of reliable foreground evidence when cross-modal agreement is available, thereby serving as an effective anchor for target perception. However, consensus support alone is insufficient for complete foreground awareness. As RGB quality deteriorates, the GT-region consensus response decreases, while the discrepancy response correspondingly increases and eventually becomes dominant. This transition suggests that accurate perception of the true foreground requires not only agreement-based evidence, but also discrepancy-aware compensation when cross-modal consistency weakens.

A seemingly counter-intuitive phenomenon is that the combined GT-region responses of consensus and discrepancy support can exceed the mean foreground occupancy ratio. This is nevertheless expected, because foreground occupancy measures geometric area proportion, whereas NMRP measures response magnitude. For the same pixel, both consensus and discrepancy branches can produce nonzero values; the difference in their magnitudes reflects the branch preference of that pixel, while both responses together still characterize its full discriminative attributes. Therefore, the sum of the two responses should not be interpreted as a partition of foreground area. Instead, it reflects the complementary encoding of target evidence by the two support patterns, which further explains why both consensus and discrepancy modeling are necessary for robust multispectral perception.

\end{document}